\documentclass[journal]{IEEEtran}
\usepackage[colorinlistoftodos]{todonotes}
\usepackage [colorlinks=true,linkcolor=blue]{hyperref}
\usepackage[utf8]{inputenc} 
\usepackage[T1]{fontenc} 
\usepackage[square,sort,comma,numbers]{natbib}
\usepackage{amsmath, amsfonts} 
\usepackage{dsfont}
\usepackage{lipsum}
\usepackage{breqn}
\usepackage{amsmath}
\usepackage{amssymb}
\usepackage{color}
\usepackage{rotating}
\usepackage{bigstrut}
\usepackage{multirow}
\usepackage{multicol}
\usepackage{tabularx}
\usepackage{colortbl}
\usepackage{hhline}
\usepackage{tabularx}
\usepackage{graphbox}
\usepackage{graphicx} 
\usepackage{booktabs} 
\usepackage{xparse}   
\usepackage{dsfont}
\usepackage{nth}
\usepackage{bbm}
\usepackage{xparse}   
\usepackage{xspace}
\usepackage{gensymb}
\usepackage{enumitem}
\usepackage[utf8]{inputenc}
\usepackage{graphicx}
\usepackage{grffile}
\usepackage[export]{adjustbox}
\usepackage{footnote}
\usepackage{url}

\usepackage{gensymb}
\usepackage{enumitem}
\usepackage[utf8]{inputenc}
\usepackage{graphicx}
\usepackage{grffile}
\usepackage[export]{adjustbox}
\usepackage{footnote}
\usepackage{url}
\usepackage[font={small,it}]{caption}
\usepackage{subcaption}
\usepackage{hyperref}
\usepackage[capitalise,noabbrev]{cleveref}

\usepackage{pgfplots}
\pgfplotsset{compat=newest}
\usepackage{tabularx}

\usepackage{hyperref}
\newcolumntype{Y}{>{\raggedright\arraybackslash}X}

\graphicspath{ {figures/} }

\newcommand\st{\textsuperscript{st}\xspace}
\newcommand\nd{\textsuperscript{nd}\xspace}
\newcommand\rd{\textsuperscript{rd}\xspace}
\newcommand\nnth{\textsuperscript{th}\xspace} 

\usepackage{scrextend}
\usepackage{changepage}
\usepackage{array}
\usepackage{booktabs}
\usepackage{dcolumn}
\newcolumntype{x}{>{\hspace{3pt}}l<{\hspace{3pt}}}
\newcolumntype{X}{>{\hspace{3pt}\scriptsize}l<{\hspace{3pt}}}
\newcolumntype{o}[1]{>{\hspace{3pt}\centering}p{#1}}
\newcolumntype{O}{>{\scriptsize}c}
\newcolumntype{v}[1]{\hspace{3pt}p{#1}}
\newcolumntype{V}[1]{>{\scriptsize\raggedright\hspace{0pt}}p{#1}}
\newcolumntype{n}{l}
\newcolumntype{N}{>{\scriptsize}l}
\usepackage{siunitx}
\DeclareSIUnit\px{px}
\usepackage{glossaries}
\newacronym[]{gls:RS}{RS}{remote sensing}
\newacronym[]{gls:GSD}{GSD}{ground sampling distance}
\newacronym[]{gls:CNN}{CNN}{convolutional neural network}
\newacronym[]{gls:CRF}{CRF}{conditional random field}
\newacronym[]{gls:SGD}{SGD}{Stochastic Gradient Descent}
\newacronym[]{gls:FCNN}{FCNN}{fully convolutional neural network}
\newacronym[]{gls:DCNN}{DCNN}{deep convolutional neural network}
\newacronym[]{gls:ACNN}{ACNN}{atrous convolutional neural network}
\newacronym[longplural={oriented bounding boxes}]{gls:OBB}{OBB}{oriented bounding box}
\newacronym[longplural={horizontal bounding boxes}]{gls:HBB}{HBB}{horizontal bounding box}
\newacronym[]{gls:RCNN}{RCNN}{Region-based convolutional neural network}

\newacronym[]{gls:miou}{mIoU}{mean intersection over union}

\newacronym[]{AV}{AV}{autonomous vehicles}
\newacronym[]{HD}{HD}{high definition}
\newacronym[]{ADAS}{ADAS}{advanced vehicle assistance system}
\newacronym[]{GPS}{GPS}{global positioning system}
\newacronym[]{USGS}{USGS}{U.S. Geological Survey}
\newacronym[]{DWT}{DWT}{discrete wavelet transform}
\newacronym[]{SVM}{SVM}{support vector machine}
\newacronym[]{DSM}{DSM}{digital surface model}
\newacronym[]{IoU}{IoU}{intersection over union}

\newglossaryentry{gls:FastRCNN}{name={Fast-\gls{gls:RCNN}}, description={}}
\newglossaryentry{gls:FasterRCNN}{name={Faster-\gls{gls:RCNN}}, description={}}
\newglossaryentry{gls:MaskRCNN}{name={Mask-\gls{gls:RCNN}}, description={}}
\glsdisablehyper  
\defglsentryfmt{
	\ifglsused{\glslabel}{%
		\glsgenentryfmt%
	}{%
	\emph{\glsgenentryfmt}%
}%
}
\usepackage[T1]{fontenc}
\usepackage{selinput}
\usepackage{mathtools}

\usepackage{tumabbrev}
\newcommand{\goal}{%
\begin{figure}[t] 
\centering
\includegraphics[width=0.5\textwidth]{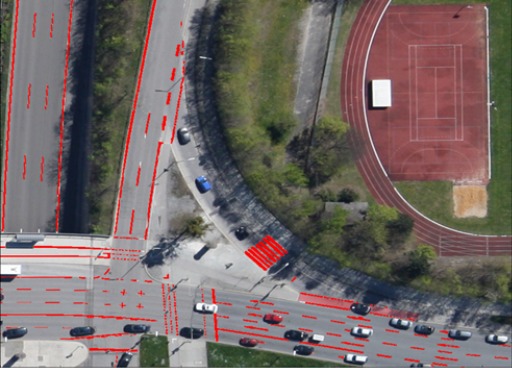}
\caption{Sample aerial image patch from AerialLanes18 dataset in which lane markings have been annotated. In this task, all classes of lane markings have been considered for pixel-wise semantic segmentation.}
\label{fig:goal}
\end{figure}
}

\newcommand{\challenges}{%
\begin{figure}[t]
\centering
\includegraphics[width=0.5\textwidth]{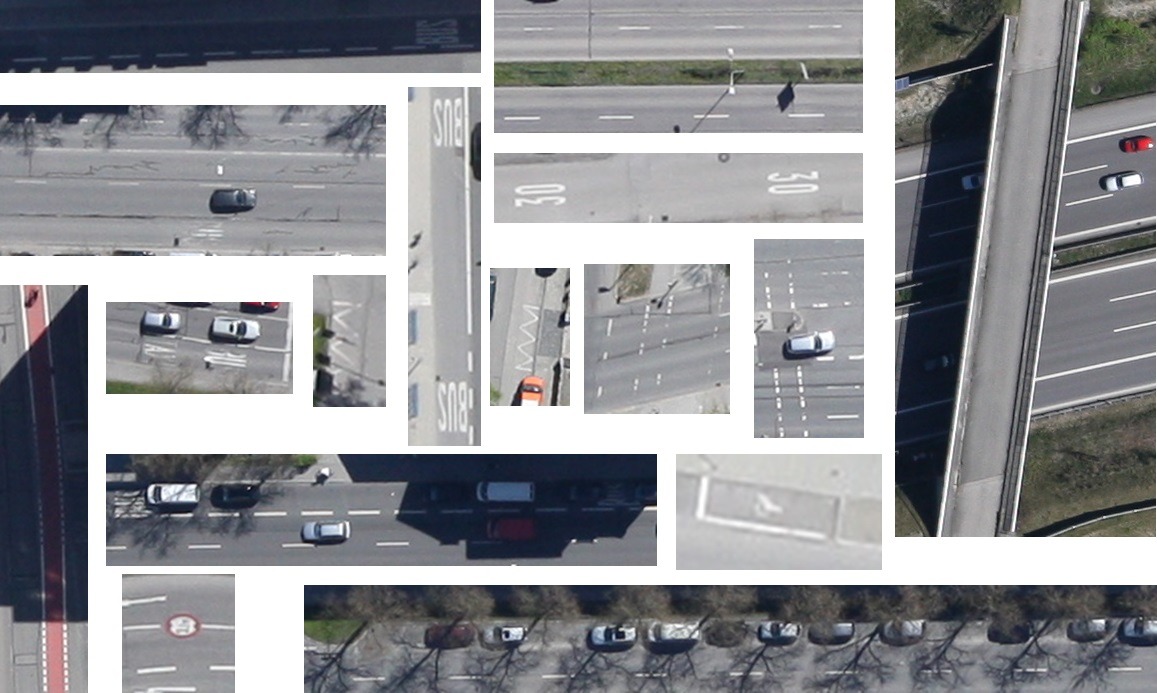}
\caption{Challenges in lane marking segmentation. Light and strong Shadow caused by trees and buildings. Examples of rare cases such as speed limit and the disabled, and bus signs have been indicated. Partial or total occlusion by other objects such as bridge or tree branches can be seen.}
\label{fig:challenges}
\end{figure}
}

\newcommand{\classes}{%
\begin{figure}[t] 
\centering
\includegraphics[width=0.5\textwidth]{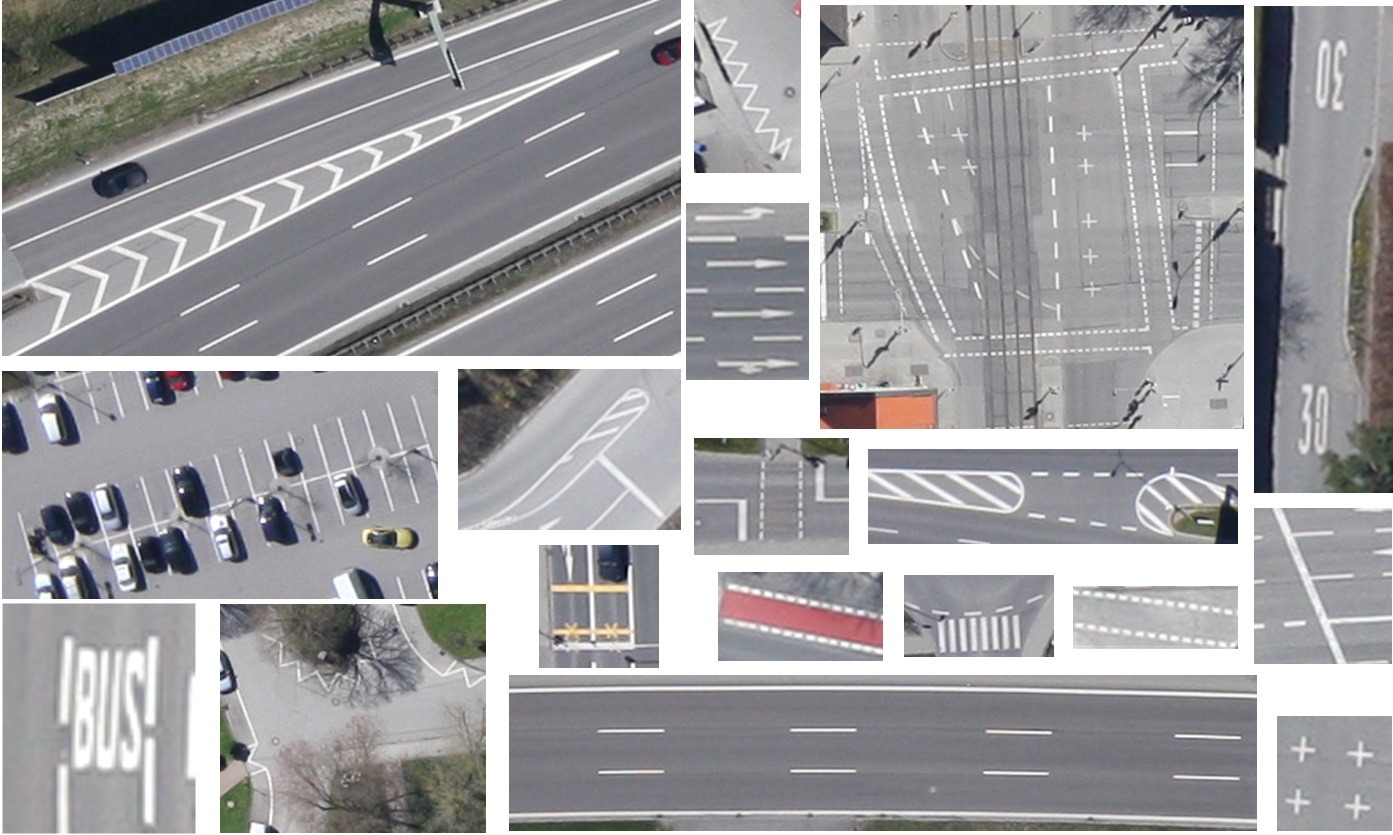}
\caption{Different lane marking classes. Single and double boundary, intersection, boxed junction, turn signs, separator, zig-zag, bus and bike sign, speed limit, no-parking zone and pedestrian crossing.}
\label{fig:classes}
\end{figure}
}

\newcommand{\similarobjects}{%
\begin{figure}[t]
\centering
\includegraphics[width=0.5\textwidth]{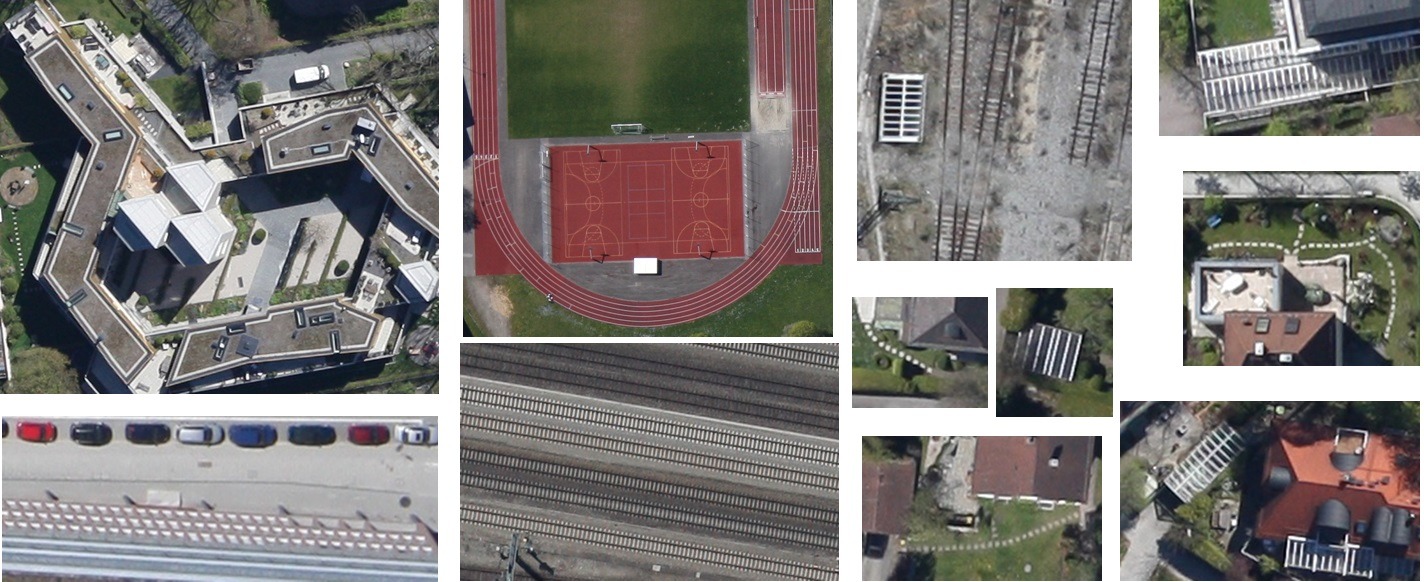}
\caption{Complex Background. Objects such as those shown in this figure share similar appearance with lane markings. As some complex background cases one can name sport field lines, rail ways, roofs of buildings and so on.}
\label{fig:similarobjects}
\end{figure}
}

\newcommand{\overview}{%
\begin{figure*}[t]	
\centering
\includegraphics[width=\textwidth]{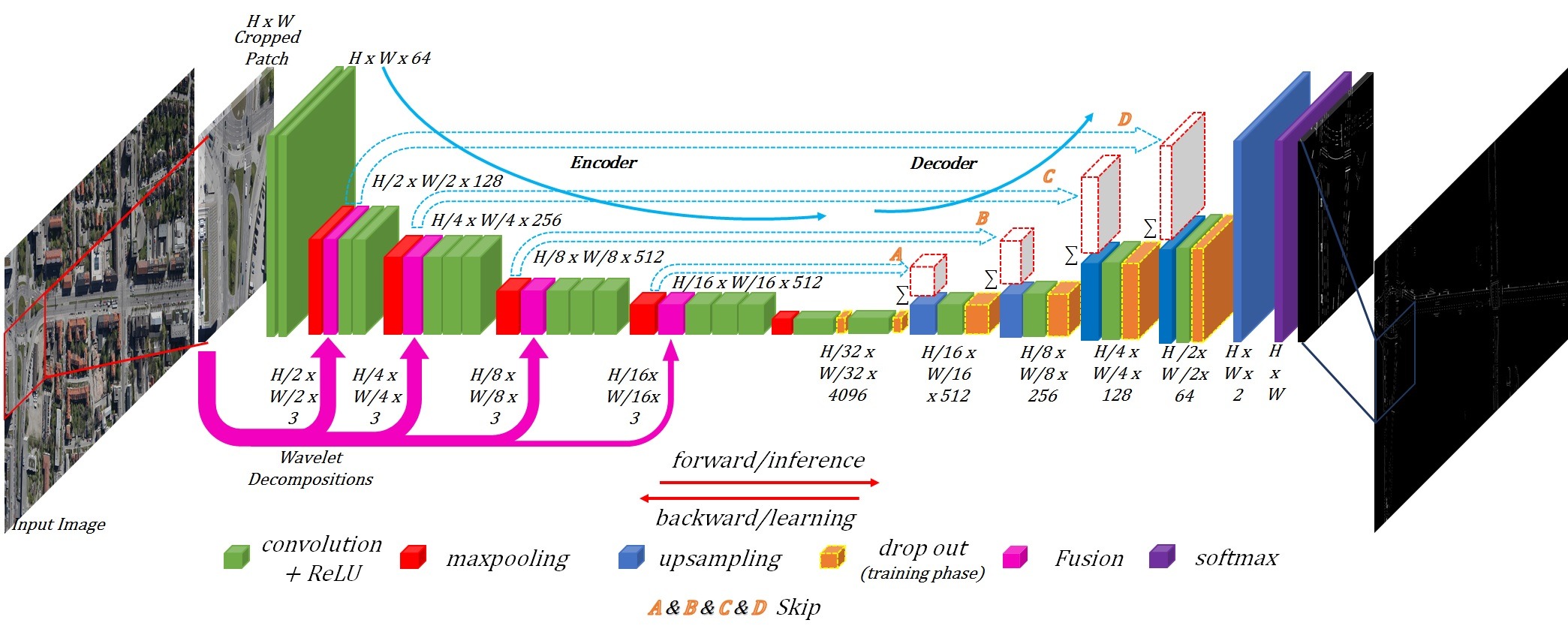}
\vspace{-0.7cm}
\caption{Aerial LaneNet. Overview of lane marking segmentation approach using Wavelet-enhanced symmetric cost-sensitive fully convolutional neural networks. The input image is a high resolution aerial image. It is cropped first and segmented using Aerial LaneNet network. In the end, segmented patches are stitched together. H and W represent height and width and third number is number of feature maps.}
\vspace{-0.4cm}
\label{fig:overview}
\end{figure*}
}

\newcommand{\waveletone}{%
\begin{figure}[t]
\centering
\includegraphics[width=0.5\textwidth]{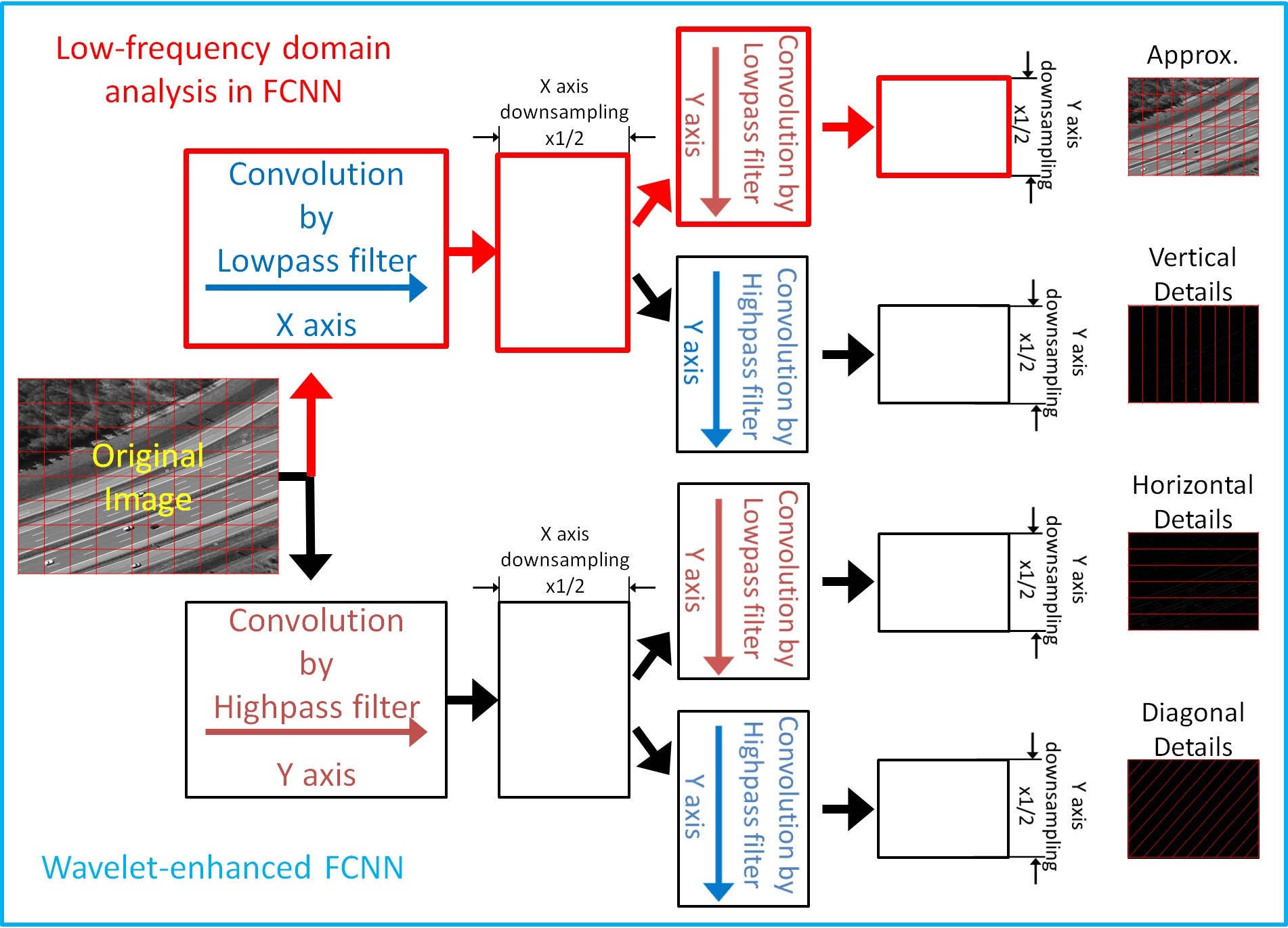}
\caption{First level DWT decomposition work flow. The input gray-scale image is processed by low pass and high pass filter in different directions. The output is with half size of the original image. Afterwards, the same operation is applied on each part, resulting in 4 decomposition parts of the input image in 1st level DWT. In conventional FCNNs, only low-frequency analysis is carried out shown in red, while DWT offers a full spectral analysis shown in blue.}
\label{fig:wavelet1}
\end{figure}
}

\newcommand{\wavelettwo}{%
\begin{figure}[t]
\centering
\includegraphics[width=0.5\textwidth]{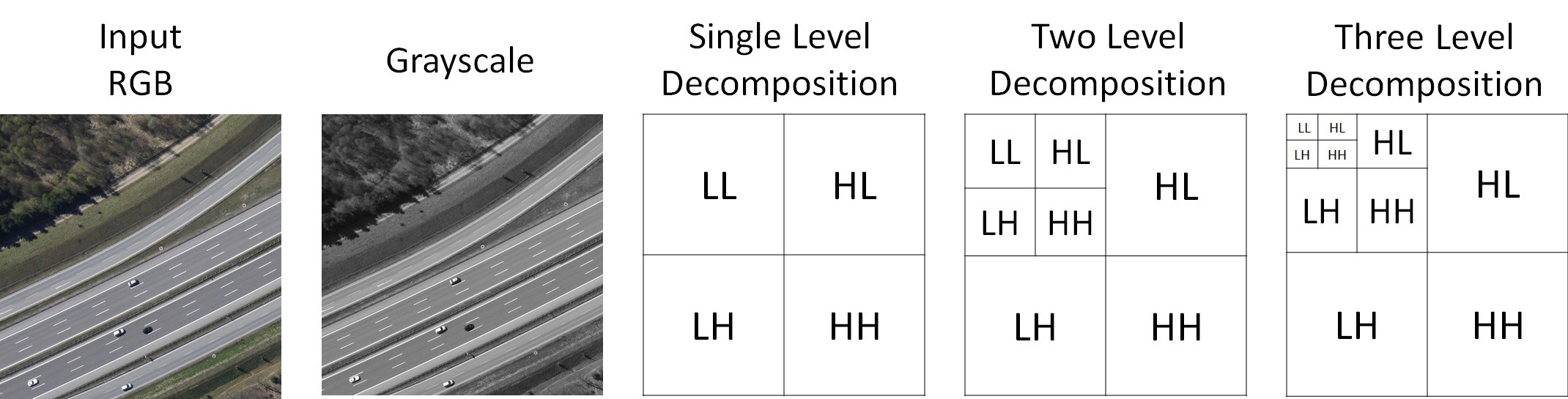}
\caption{Different DWT decompositions. The input RGB image is converted to gray scale first. Then first DWT decomposition is computed followed by next levels. High-pass and low-pass filters are represented by $``H''$ and $``L''$ respectively. LL stands for two step low-pass filtering where HL, LH and HH contain horizontal, vertical and diagonal details respectively.}
\label{fig:wavelet2}
\end{figure}
}
\newcommand{\differentfusion}{%
\begin{figure*}
\centering
\includegraphics[width=1\textwidth]{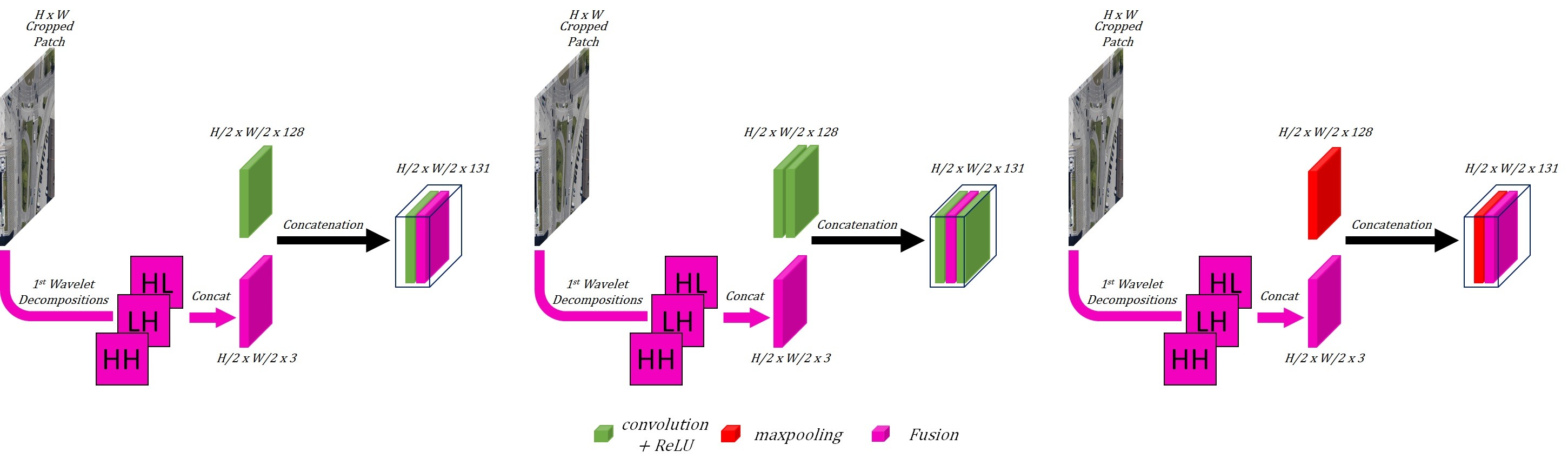}
\caption{Different 1st level DWT fusion with Symmetric FCNNs. There are three fusion variants. Left: before pooling layer, middle: after convolution layer, right: after pooling layer.}
\label{fig:differentfusion}
\end{figure*}
}
\newcommand{\breakdown}{%
\begin{figure*}[t]
\centering
\includegraphics[width=1\textwidth]{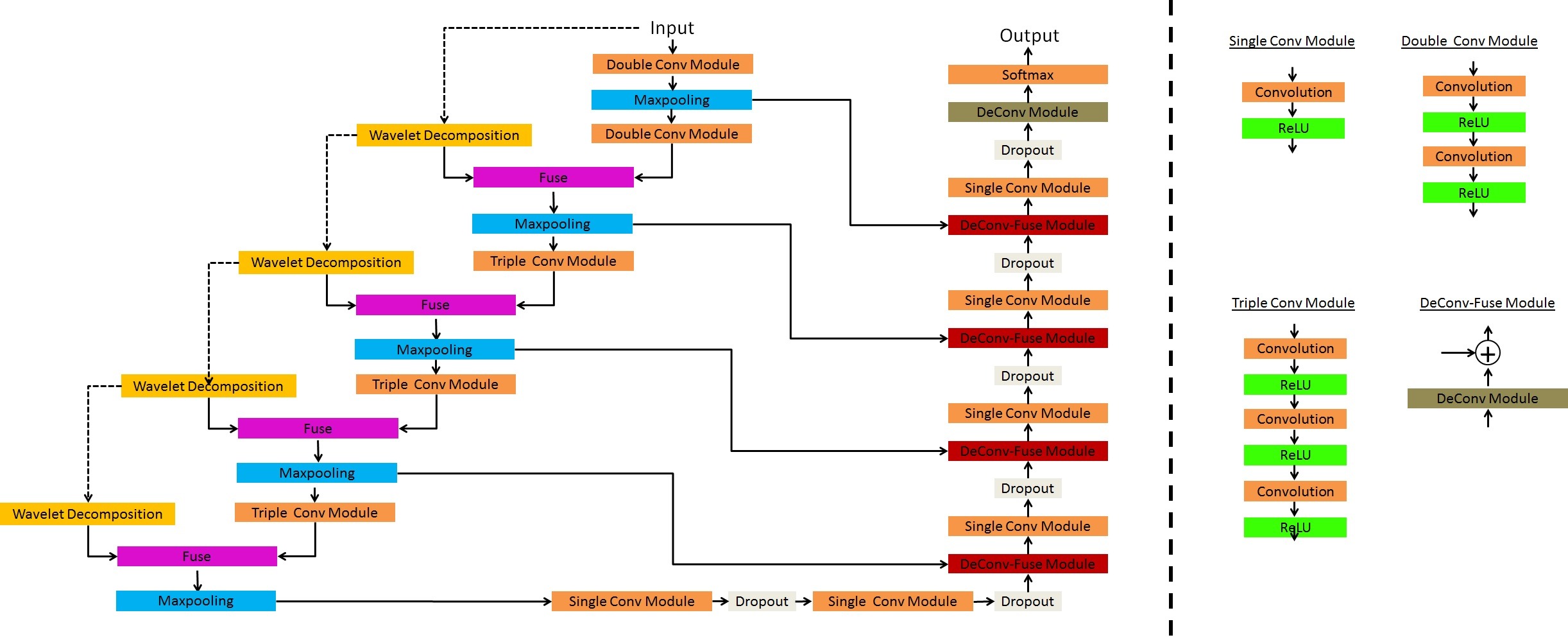}
\caption{Aerial LaneNet architecture break down.}
\label{fig:breakdown}
\end{figure*}
}

\newcommand{\patchtrimgone}{%
\begin{figure*}
\includegraphics[width=1\linewidth]{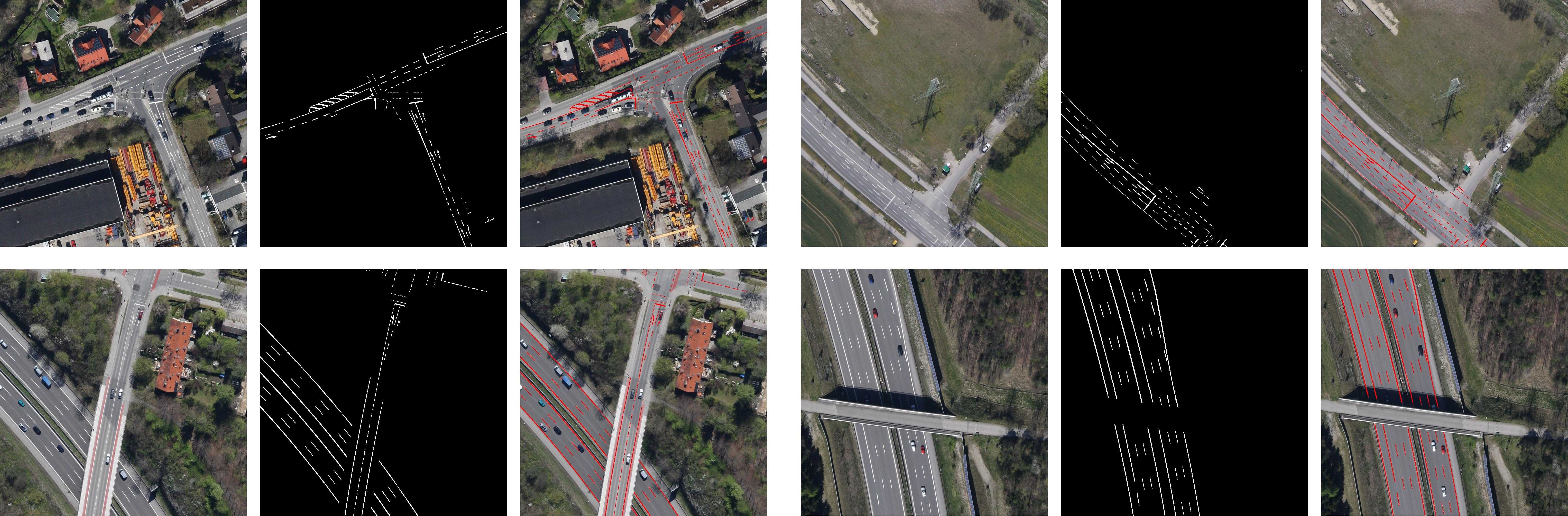}
\caption{Sample training patches from AerialLanes18 dataset taken by aerial imagery over Munich, Germany. The original image patch is shown with its corresponding annotation. GSD is 13cm.}
\label{fig:patchtrimg1}
\end{figure*}
}

\newcommand{\wholetrimgone}{%
\begin{figure*}
\begin{subfigure}{1\textwidth}
\includegraphics[width=1\linewidth]{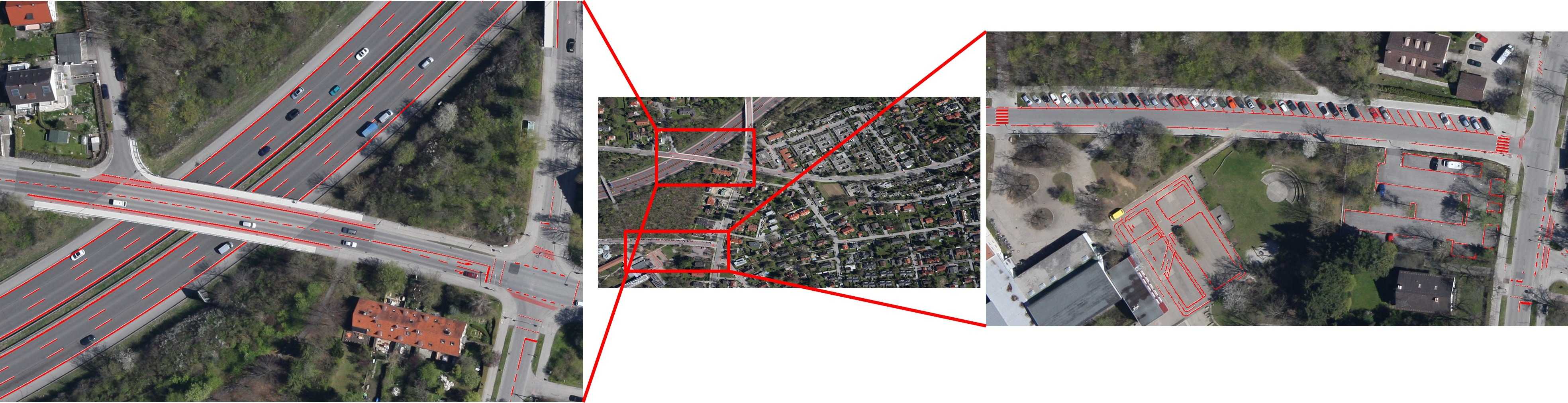}
\end{subfigure}
\begin{subfigure}{1\textwidth}
\includegraphics[width=1\linewidth]{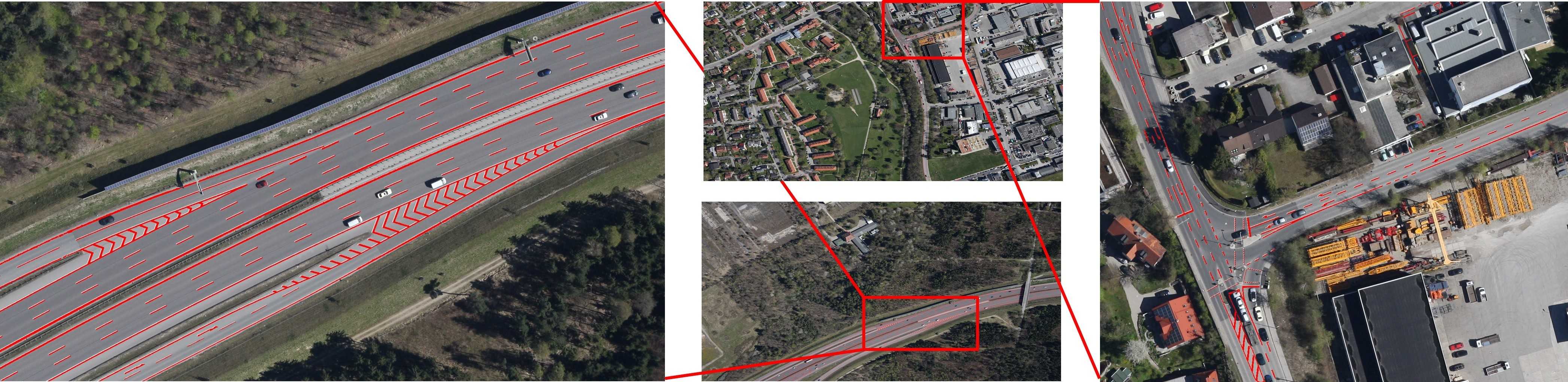}
\end{subfigure}
\caption{Sample large training image from AerialLanes18 dataset. The original image patch is shown with its corresponding annotation.}
\label{fig:wholetrimg1}
\end{figure*}
}

\newcommand{\lambdaEffect}{%
\begin{figure}
\begin{tikzpicture}
\begin{semilogxaxis}[
xlabel=$\lambda_{lane}$,
ylabel=mean IoU]
\addplot[smooth, thick, color = red, mark = *] coordinates {
(1,55.23)
(50,56.77)
(100,57.22)
(200,57.93)
(300,58.12)
(308,58.2)
(350,58.45)
(400,59.06)
(500,58.76)
(1000,57.32)
};
\end{semilogxaxis}
\end{tikzpicture}
\caption{Performance of FCN32s network with AlexNet as backbone network on different $\lambda_{lane}$ values during training. The ratio between lane marking and background pixels in train, trainval and test set are 389, 418 and 308 respectively.}
\label{fig:lambda}
\end{figure}
}

\newcommand{\eachimageone}{
\begin{figure}[t]
\begin{tikzpicture}
\begin{axis}[
x tick label style={
/pgf/number format/1000 sep=},
ylabel=\%percentage,
xlabel=\# image number,
enlargelimits=0.05,
ymax=100,
legend style={
legend columns=-1},
ybar interval=0.7 
]
\addplot 
coordinates {(1,80.48) (2,81.02) (3,80.69) (4,91.86) (5,88.01) (6,83.85) (7,83.54) (8,94.88) (9,87.86) (10,87.29) (11,77.8)};

\addplot 
coordinates {(1,85.53) (2,86.48) (3,86.04) (4,84.13) (5,83.92) (6,78.2) (7,85.68) (8,94.14) (9,86.43) (10,83.8) (11,77.8)};

\legend{recall,precision}
\end{axis}
\end{tikzpicture}
\caption{Evaluation of Aerial LaneNet network with total recall and precision values for each test image.}
\label{fig:eachimage1}
\end{figure}
}

\newcommand{\eachimagetwo}{
\begin{figure}[t]
\begin{tikzpicture}
\begin{axis}[
x tick label style={
/pgf/number format/1000 sep=},
ylabel=\%percentage,
xlabel=\# image number,
enlargelimits=0,
legend style={
legend columns=3},
ybar interval=0.7 
]
\addplot 
coordinates {(1,74.34) (2,	75.2) (3,	74.69) (4,	80.12) (5,	77.92) (6,	72.17) (7,	76.39) (8,	90.09) (9,	79.49) (10,	77.43) (11,55.0)};

\addplot 
coordinates {(1,65.78) (2,	67.13) (3,	66.51) (4,	75.23) (5,	71.74) (6,	61.62) (7,	69.21) (8,	89.05) (9,	74.38) (10,	71.01) (11,55.0)};
    
\addplot 
coordinates {(1,61.08) (2,	62.1) (3,	61.54) (4,	83.77) (5,	76.08) (6,	67.8) (7,	67.13) (8,	89.79) (9,	75.85) (10,	74.69) (11,55.0)};

\addplot 
coordinates {(1,99.87) (2,	99.95) (3,	99.83) (4,	99.96) (5,	99.94) (6,	99.89) (7,	99.94) (8,	99.96) (9,	99.88) (10,	99.88) (11,55.0)};
    
\addplot 
coordinates {(1,71.27) (2,	73.05) (3,	72.35) (4,	68.27) (5,	67.87) (6,	56.46) (7,	71.43) (8,	88.32) (9,	72.97) (10,	67.68) (11,55.0)};

\addplot 
coordinates {(1,99.79) (2,	99.92) (3,	99.73) (4,	99.98) (5,	99.96) (6,	99.93) (7,	99.93) (8,	99.97) (9,	99.9) (10,	99.92) (11,55.0)};

\legend{mean \gls{IoU}, dice, recall lane, recall bg, precision lane, precision bg}
\end{axis}
\end{tikzpicture}
\caption{Evaluation of Aerial LaneNet network on each test image with mean \gls{IoU}, dice and recall and precision values for each class.}
\label{fig:eachimage2}
\end{figure}
}

\newcommand{\sFigure}{
\begin{figure*}
\begin{subfigure}{1\textwidth}
\centering
\includegraphics[width=\linewidth]{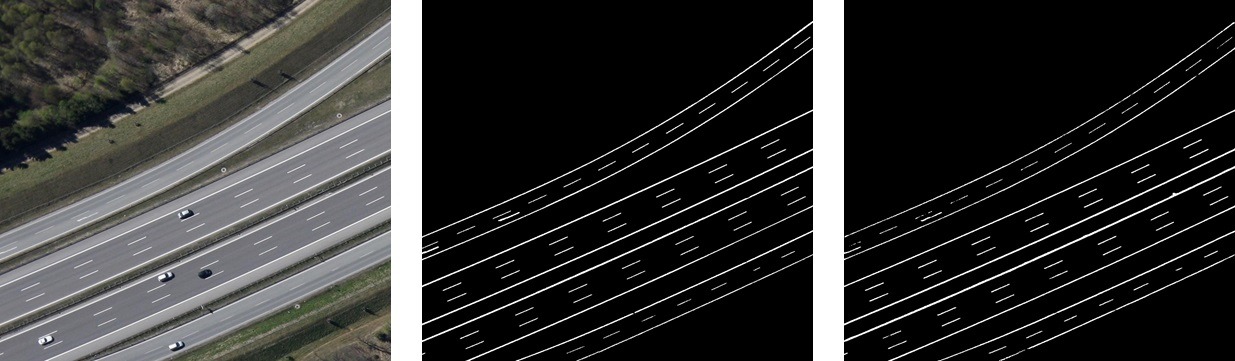}
\end{subfigure}
\begin{subfigure}{1\textwidth}
\centering
\vspace{0.1cm}
\includegraphics[width=\linewidth]{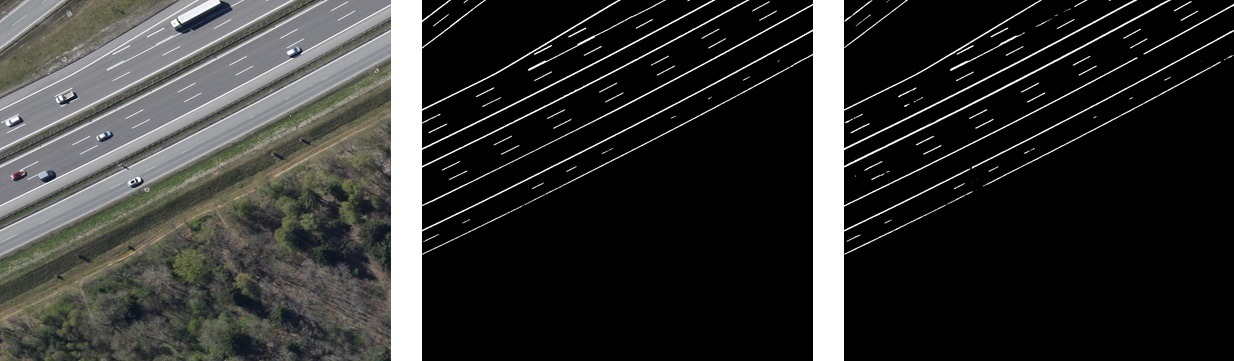}
\label{fig:highway}
\end{subfigure}
\caption{Examples of results using Aerial LaneNet approach with the best performance. The left column shows input images. The middle columns shows ground truth and the right column images are predictions.}
\label{fig:s}
\end{figure*}
}

\newcommand{\BenchmarkFigure}{
\begin{figure*}
\begin{adjustwidth}{-0.5cm}{0cm}
\begin{subfigure}[t]{0.2\textwidth}
\includegraphics[width=\linewidth]{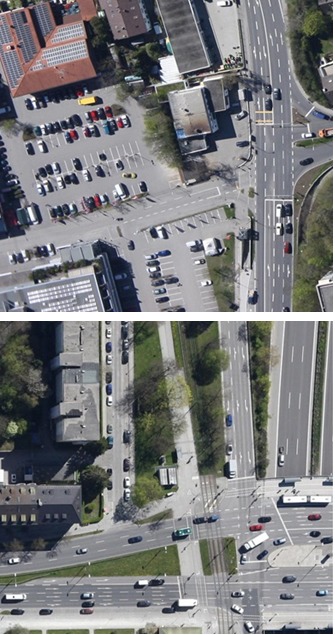}
\caption{input patch}
\end{subfigure}
\begin{subfigure}[t]{0.2\textwidth}
\includegraphics[width=\linewidth]{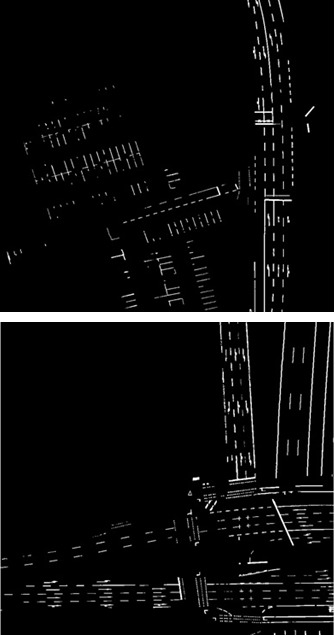}
\caption{ground truth}
\end{subfigure}
\begin{subfigure}[t]{0.2\textwidth}
\includegraphics[width=\linewidth]{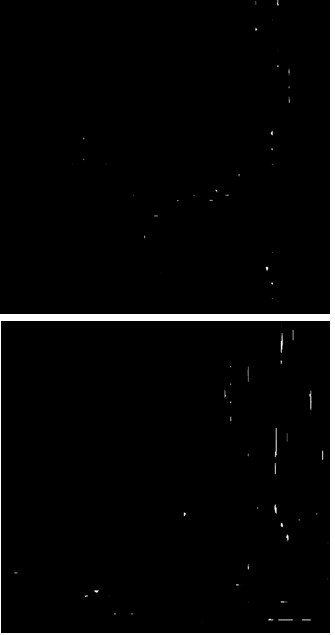}
\caption{DeepLabv3}
\end{subfigure}
\begin{subfigure}[t]{0.2\textwidth}
\includegraphics[width=\linewidth]{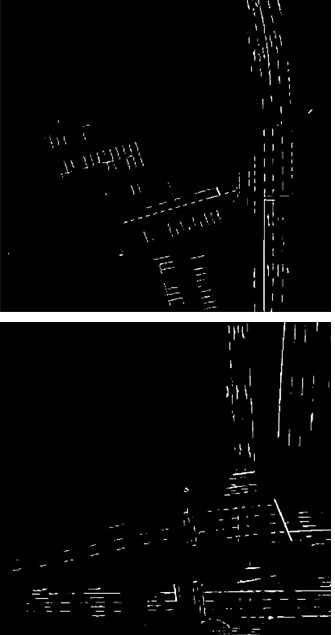}
\caption{DeepLabv3+}
\end{subfigure}
\begin{subfigure}[t]{0.2\textwidth}
\includegraphics[width=\linewidth]{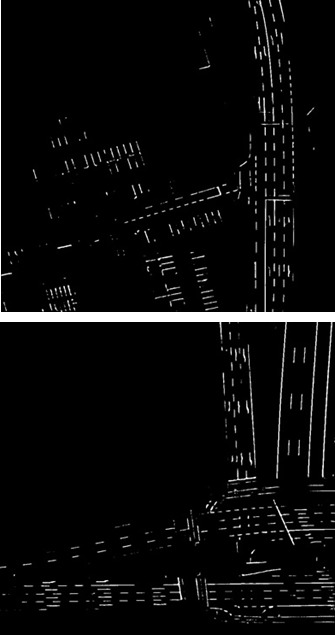}
\caption{Aerial LaneNet}
\end{subfigure}
\caption{Qualitative comparison of Aerial LaneNet with ground truth and the state-of-the-art algorithms DeepLabv3 and DeepLabv3+.}
\label{fig:qualitativebenchmark}
\end{adjustwidth}
\end{figure*}
}

\newcommand{\overlaidorgone}{
\begin{figure*}[t]
\centering
\includegraphics[width=1\linewidth]{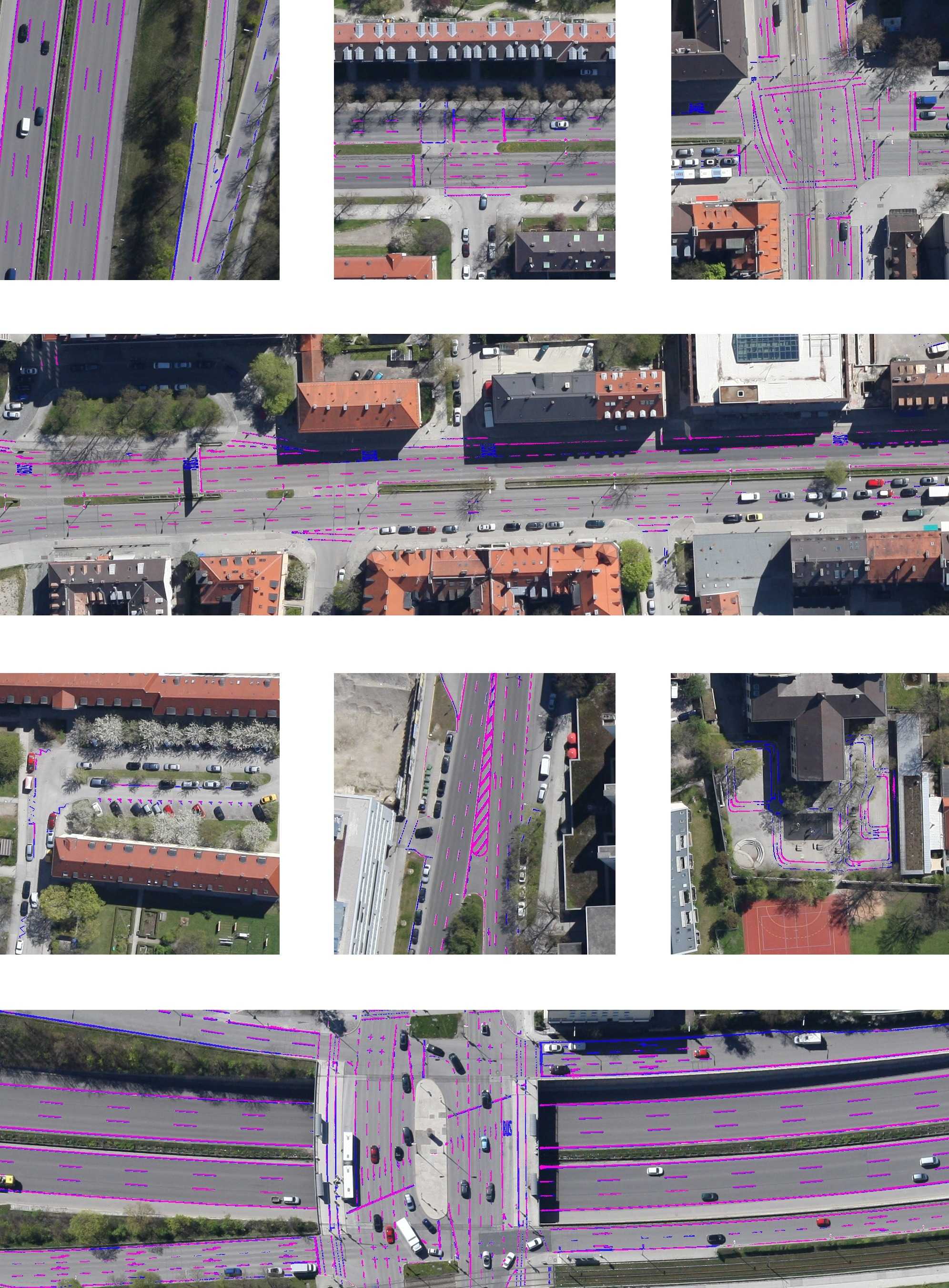}
\caption{Test image with overlaid prediction and ground truth. Ground truth which has not been predicted has been illustrated with dark blue color and prediction is depicted with pink color.}
\label{fig:overlaidorg1}
\end{figure*}
}

\newcommand{\newtestpatch}{
\begin{figure*}[t]
\centering
\includegraphics[width=1\textwidth]{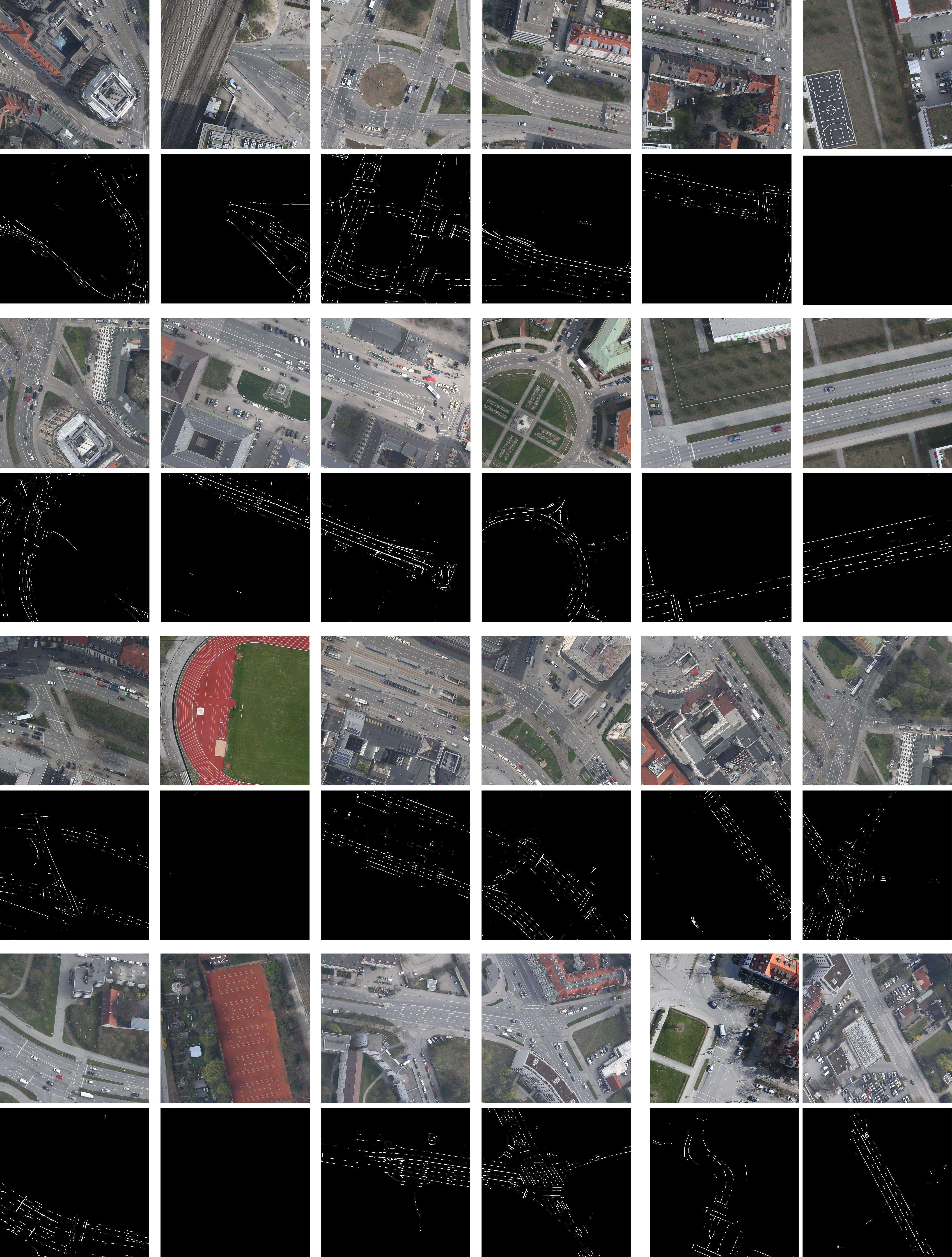}
\caption{New test patch images taken in different days, GSD and camera angles in comparison with AerialLanes18 dataset. Each patch has been shown with the corresponding lane marking segmentation.}
\label{fig:newtestpatch}
\end{figure*}
}

\newcommand{\newtestpatchoverlaid}{
\begin{figure*}[t]
\centering
\includegraphics[width=\textwidth]{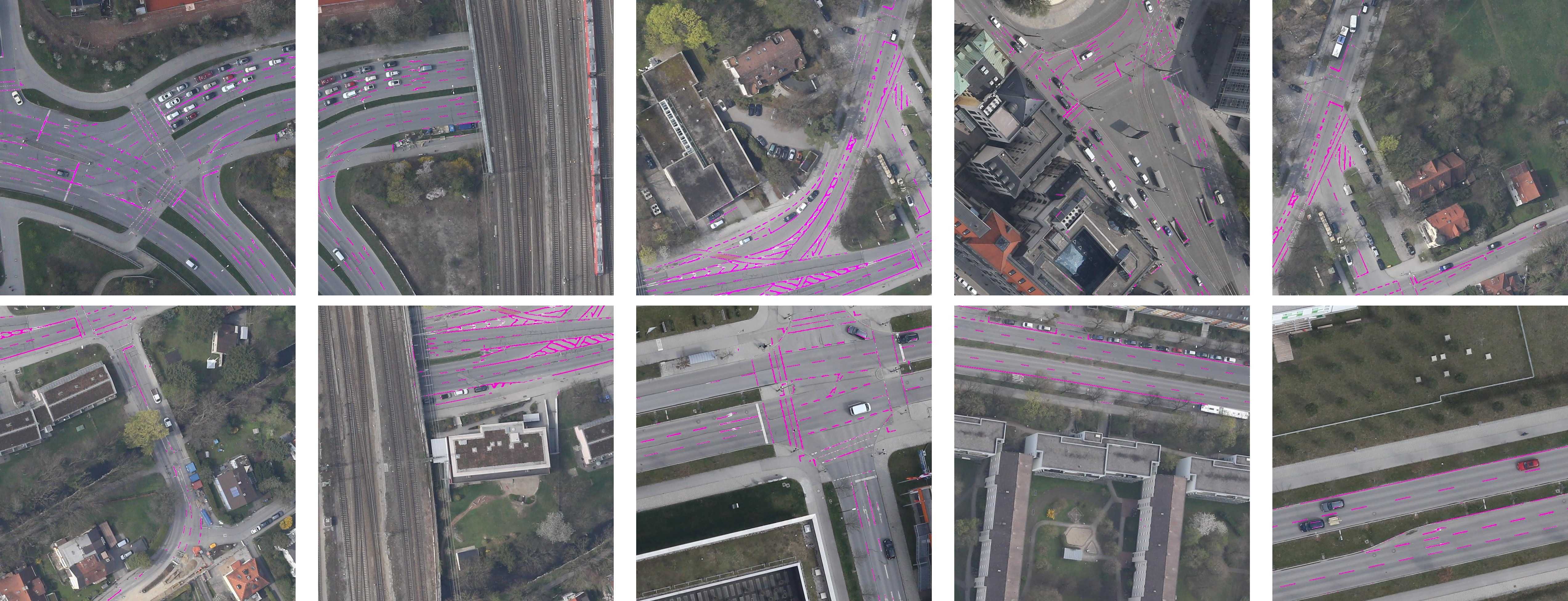}
\caption{New test patch images taken in different days, GSD and camera angles in comparison with AerialLanes18 dataset. Lane Marking prediction has been overlaid on patches in order to illustrate the localization accuracy of Aerial LaneNet network.}
\label{fig:newtestpatchoverlaid}
\end{figure*}
}
\newcommand{\receptivefieldsTable}{%
\begin{table*}
  \centering
 \caption{Symmetric FCNN input and output sizes for each layer as well as filter maps and receptive fields. Added layers in Symmetric FCNN to FCN8s have been specified with red colors.}
\begin{tabular}{ |c|c|c|c|c|c| } 
\hline
        Layer&
        Input&
        Output&
        Features&
        Receptive Field \\
\hline
\hline
conv1-1&$960\times960\times3$&$960\times960\times64$&64&$3\times3$\\
\hline
conv1-2&$960\times960\times64$&$960\times96\times64$&64&$5\times5$\\

\hline
maxpooling-1/conv2-1&$960\times960\times64$&$480\times480\times128$&128&$11\times11$\\
\hline
conv2-2/1\st level Wavelet-fusion&$480\times480\times128$&$480\times480\times131$&131&$13\times13$\\

\hline
maxpooling-2/conv3-1&$480\times480\times131$&$240\times240\times256$&256&$17\times17$\\
\hline
conv3-2&$240\times240\times256$&$240\times240\times256$&256&$19\times19$\\
\hline
conv3-3/2\nd level Wavelet-fusion&$240\times240\times256$&$240\times240\times259$&259&$21\times21$\\
\hline

maxpooling-3/conv4-1&$240\times240\times256$&$120\times120\times512$&512&$25\times25$\\
\hline
conv4-2&$120\times120\times51$2&$120\times120\times512$&512&$27\times27$\\
\hline
conv4-3/3\rd level Wavelet-fusion&$120\times120\times512$&$120\times120\times515$&$515$&$29\times29$\\
\hline
maxpooling-4/conv5-1&$120\times120\times515$&$60\times60\times512$&$512$&$33\times33$\\
\hline
conv5-2&$60\times60\times512$&$60\times60\times512$&$512$&$35\times35$\\
\hline
conv5-3/4\nnth level Wavelet-fusion&$60\times60\times512$&$60\times60\times515$&$515$&$37\times37$\\
\hline

maxpooling-5/conv6-1&$60\times60\times515$&$30\times30\times4096$&$4096$&$41\times41$\\
\hline
dropout-1&-&-&-&-\\
\hline
conv6-2&$30\times30\times4096$&$30\times30\times4096$&$4096$&$43\times43$\\
\hline
dropout-2&-&-&-&-\\
\hline
deconv-1/maxpooling-1-fusion&$30\times30\times4096$&$60\times60\times512$&$512$&$43\times43$\\
\hline
\textcolor{red}{conv7}&$60\times60\times512$&$60\times60\times512$&$512$&$43\times43$\\
\hline
\textcolor{red}{dropout-3}&-&-&-&-\\
\hline

deconv-2/maxpooling-2-fusion&$60\times60\times512$&$120\times120\times256$&256&$43\times43$\\
\hline
\textcolor{red}{conv8}&$120\times120\times256$&$120\times120\times256$&256&$43\times43$\\
\hline
\textcolor{red}{dropout-4}&-&-&-&-\\
\hline

deconv-3/maxspooling-3-fusion&$120\times120\times256$&$240\times240\times128$&128&$43\times43$\\
\hline
\textcolor{red}{conv9}&$240\times240\times128$&$240\times240\times128$&128&$43\times43$\\
\hline
\textcolor{red}{dropout-5}&-&-&-&-\\
\hline
deconv-4/maxpooling-4-fusion&$240\times240\times128$&$480\times480\times64$&64&$43\times43$\\
\hline
\textcolor{red}{conv10}&$480\times480\times64$&$480\times480\times64$&64&$43\times43$\\
\hline
\textcolor{red}{dropout-6}&-&-&-&-\\
\hline
deconv-5&$480\times480\times64$&$960\times960\times2$&2&$43\times43$\\

\hline
\end{tabular}
\label{tab:receptivefields}
\end{table*}
}

\newcommand{\2}{%
\begin{table*}
    \centering
    \caption{Evaluation of lane marking segmentation using different backbone networks for segmentation with one up-sampling layer.
    With VGG16 network, this is equivalent with FCN32s.
    \change{In fine-tuning, the parameters are initialized by ImageNet pre-trained model rather than random initialization.
    In this case, all of the layers are re-trained.}
    Mean IoU numbers in $[\%]$.
    Higher value is better. Max stride is 32pixel.}
    \label{tab:2}
    \begin{tabular}{ccccccccc}
    \toprule
        Network&
        \begin{tabular}[c]{@{}c@{}}weighted\\loss\end{tabular}&
        \begin{tabular}[c]{@{}c@{}}fine\\tuned\end{tabular}&
        \begin{tabular}[c]{@{}c@{}}data\\augmentation\end{tabular}&
        \begin{tabular}[c]{@{}c@{}}mean\\IoU\end{tabular}&
        \begin{tabular}[c]{@{}c@{}}forward\\time\end{tabular}&
        \begin{tabular}[c]{@{}c@{}}conv.\\layers\end{tabular}&
        param.
        \\\midrule
        FCN-AlexNet~\cite{Long2015FullySegmentation}&-&-&-&51.08&80ms&8&57M\\
        FCN-AlexNet&-&-&\checkmark&52.92&80ms&8&57M\\
        FCN-AlexNet&-&\checkmark&\checkmark&55.23&80ms&8&57M\\
        FCN-AlexNet&\checkmark&\checkmark&\checkmark&59.06&80ms&8&57M\\
        FCN-VGG16~\cite{Long2015FullySegmentation}&\checkmark&\checkmark&\checkmark&\textbf{61.56}&300ms&16&134M\\
        FCN-GoogLeNet~\cite{Long2015FullySegmentation}&\checkmark&\checkmark&\checkmark&61.49&100ms&22&6M\\
        \bottomrule
    \end{tabular}
\end{table*}
}

\newcommand{\lambdaTable}{
\begin{table*}
    \centering
    \caption{Numerical results of FCN32s-AlexNet using different values of $\lambda_{lane}$ during training.
    The base network is VGG16.}
    \label{tab:lambda}
    \begin{tabular}{ccccccccccc}
    \toprule
        $\lambda_{lane}$ value &1&50&100&200&300&308&350&400&500&1000\\\midrule
        mean IoU                &55.23&56.77&57.22&57.93&58.12&58.21&58.45&\textbf{59.06}&58.76&57.32\\
        \bottomrule
    \end{tabular}
\end{table*}
}

\newcommand{\symmetric}{
\begin{table}
    \centering
    \caption{Impact of added convolutions, drop-out and up-sampling layers to shape Symmetric FCNN on AerialLanes18 dataset.
    The base network is VGG16.}
    \label{tab:symmetric}
    \begin{tabular}{ccccccccc}
    \toprule
        Network&
        \begin{tabular}[c]{@{}c@{}}pixel\\acc.\end{tabular}&
        \begin{tabular}[c]{@{}c@{}}mean\\acc.\end{tabular}&
        \begin{tabular}[c]{@{}c@{}}mean\\IoU\end{tabular}&
        \begin{tabular}[c]{@{}c@{}}f.w.\\IoU\end{tabular} &
        \begin{tabular}[c]{@{}c@{}}dice s. c.\end{tabular}\\\midrule
        \begin{tabular}[c]{@{}c@{}}FCN-8s~\cite{Long2015FullySegmentation} \\(A,B and C layers)\end{tabular}&99.73&66.12&62.79&99.53&51.67\\
        \begin{tabular}[c]{@{}c@{}}FCNN\\(A,B,C and D layers)\end{tabular}&99.73&67.42&63.45&99.54&52.33\\
        \begin{tabular}[c]{@{}c@{}}FCNN\\(A,B,C,D and conv layers)\end{tabular}&99.74&68.25&64.23&99.54&53.25\\
        Symmeteric FCNN&\textbf{99.74}&\textbf{69.57}&\textbf{65.10}&\textbf{99.55}&\textbf{55.08}\\
        \bottomrule
    \end{tabular}
\end{table}
}

\newcommand{\5}{
\begin{table*}
    \centering
    \caption{Evaluation of Aerial LaneNet for fusion of each level of DWT to Symmetric FCNN with cost-sensitive loss function.
    \change{In addition, the comparison between FCN-8s~\cite{Long2015FullySegmentation} with and without 1\st level DWT is provided.}
    }
    \label{tab:5}
    \begin{tabular}{cccccccccc}
    \toprule
        Network&base network&
        \begin{tabular}[c]{@{}c@{}}pixel\\acc.\end{tabular}&
        \begin{tabular}[c]{@{}c@{}}mean\\acc.\end{tabular}&
        \begin{tabular}[c]{@{}c@{}}mean\\IoU\end{tabular}&
        \begin{tabular}[c]{@{}c@{}}f.w.\\IoU\end{tabular} &
        \begin{tabular}[c]{@{}c@{}}dice s. c.\end{tabular}\\\midrule
         \change{FCN-8s~\cite{Long2015FullySegmentation}}&VGG16&99.73&66.12&62.79&99.53&51.67\\
        \change{FCN-8s - 1\st DWT level} &VGG16&99.75&69.67&66.24&99.56&55.14\\
        Aerial LaneNet - 1\st DWT level     &VGG16        &99.77&75.86&70.16&99.60&61.23\\
        Aerial LaneNet - 1\st, 2\nd DWT level   &VGG16       &99.79&80.83&73.57&99.62&65.55\\
        Aerial LaneNet - 1\st, 2\nd, 3\rd DWT level   &VGG16    &99.80&84.32&76.72&99.65&69.61\\
        Aerial LaneNet - 1\st, 2\nd, 3\rd, 4\nnth DWT level &VGG16   &99.81&85.72&77.78&99.67&71.17\\
        \change{Aerial LaneNet - 1\st, 2\nd, 3\rd, 4\nnth DWT level}&\change{ResNet-101}&\textbf{99.81}&\textbf{85.95}&\textbf{77.98}&\textbf{99.68}&\textbf{71.42}\\
        Aerial LaneNet - 1\st, 2\nd, 3\rd, 4\nnth, 5\nnth DWT level&VGG16&99.80&84.01&76.64&99.65&70.25\\
        \bottomrule
    \end{tabular}
\end{table*}
}

\newcommand{\6}{
\begin{table*}
    \centering
    \caption{Evaluation of impact of different \gls{DWT} decompositions in 1st level on lane marking segmentation including horizontal, horizontal and vertical, horizontal, vertical and diagonal details as well as all of decompositions consisting of approximation part.
    The base network is VGG16.}
    \label{tab:6}
    \begin{tabular}{ccccccccc}
    \toprule
        Network&
        \begin{tabular}[c]{@{}c@{}}pixel\\acc.\end{tabular}&
        \begin{tabular}[c]{@{}c@{}}mean\\acc.\end{tabular}&
        \begin{tabular}[c]{@{}c@{}}mean\\IoU\end{tabular}&
        \begin{tabular}[c]{@{}c@{}}f.w.\\IoU\end{tabular} &
        \begin{tabular}[c]{@{}c@{}}dice s. c.\end{tabular}\\\midrule
        horizontal                                      &99.78&79.72&71.96&99.62&64.34\\
        horizontal and vertical                         &99.80&84.03&75.84&99.65&68.56\\
        horizontal, vertical and diagonal               &\textbf{99.81}&\textbf{85.72}&\textbf{77.78}&\textbf{99.67}&\textbf{71.17}\\
        horizontal, vertical, diagonal and approximation&99.80&83.21&76.02&99.65&69.23\\
        \bottomrule
    \end{tabular}
\end{table*}
}

\newcommand{\7}{
\begin{table}
    \centering
    \caption{Evaluation of fusion of \gls{DWT} with symmetric \gls{gls:FCNN} in different locations.
    The base network is VGG16.
    The fusion is concatenation in all cases.}
    \begin{tabular}{cc|c|c}
    \toprule
        Fusion  &After first conv&After pooling& Before pooling\\
\midrule
        mean \gls{IoU} &76.23&\textbf{77.78}&75.42\\
        \bottomrule
    \end{tabular}
    \label{tab:7}
\end{table}
}

\newcommand{\3}{
\begin{table}
    \centering
    \caption{Aerial LaneNet comarison with the state-of-the-art algorithms.
    All numbers are in $[\%]$.}
    \label{tab:3}
    \begin{tabular}{ccccccccc}
    \toprule
        Network&
        \begin{tabular}[c]{@{}c@{}}pixel\\acc.\end{tabular}&
        \begin{tabular}[c]{@{}c@{}}mean\\acc.\end{tabular}&
        \begin{tabular}[c]{@{}c@{}}mean\\IoU\end{tabular}&
        \begin{tabular}[c]{@{}c@{}}f.w.\\IoU\end{tabular} &
        \begin{tabular}[c]{@{}c@{}}dice s. c.\end{tabular}\\\midrule
        DeepLab\cite{Chen2014SemanticCRFs}&99.73&68.02&63.95&99.54&53.07\\
        UNet\cite{unet}&99.73&67.25&63.39&99.54&52.12\\
        FCN-8s~\cite{Long2015FullySegmentation}&99.73&66.12&62.79&99.53&51.67\\
        \change{DeepLabv3}~\cite{deeplabv3}&99.68&53.79&53.24&99.38&12.26\\
        \change{DeepLabv3+}~\cite{deeplabv3+}&99.79&78.23&73.18&99.62&62.71\\
        Aerial LaneNet&\textbf{99.81}&\textbf{85.95}&\textbf{77.98}&\textbf{99.68}&\textbf{71.42}\\
        \bottomrule
    \end{tabular}
\end{table}
}

\newcommand{\confusionmatrixTable}{
\begin{table}
\begin{center}
    \caption{Confusion Matrix of Aerial LaneNet with the best performance using VGG16 base network. Matrix shows the number of samples for each class predicted by the system. Due to unbalanced multi-class problem, percentage numbers for each class shows normalized recall rates. Confusion matrix shows the number of correct and wrong classified pixels along with normalized values.}
    \label{tab:confusionmatrix}
    \begin{tabular}{c|c|c|c|c|}
   \multicolumn{2}{c}{}& \multicolumn{2}{c}{\centering  \textbf{Actual Labels}}&\multicolumn{1}{c}{}\\
\cline{3-5}
 \multicolumn{2}{c|}{}&
 {Lane Marking}
&
{\begin{tabular}[c]{@{}c@{}} Background \end{tabular} }&
\textbf{Class Precision}\\
\cline{2-5}       
\multirow{5}{*}{\rotatebox{90}{ \textbf{Predicted Labels}}}&Lane Marking&
\begin{tabular}[c]{@{}c@{}}473313\\71.55\%\end{tabular}&
\begin{tabular}[c]{@{}c@{}}205431\\0.10\%\end{tabular}&
69.73\%\\
\cline{2-5}
&\begin{tabular}[c]{@{}c@{}} Background \end{tabular}&
\begin{tabular}[c]{@{}c@{}}188196\\28.45\%\end{tabular}&
\begin{tabular}[c]{@{}c@{}}209396100\\99.90\%\end{tabular}&
99.91\%\\\cline{2-5}
&\textbf{Class Recall} &71.55\% & 99.90\% & \begin{tabular}[c]{@{}c@{}}\textbf{Total Accuracy}\\mean: 85.72\%\\absolute: 99.81\%\end{tabular}\\\cline{2-5}
        \end{tabular}
    \end{center}
\end{table}
}

\newcommand{\change}[1]{\textcolor[rgb]{0,0,0}{#1}}
\newcommand\TODO[1]{}
\newcommand\majid[1]{}
\newcommand\peterf[1]{}
\newcommand\marco[1]{}
\newcommand{\peterr}[1]{}

\begin{document}

\title{Aerial LaneNet: Lane Marking Semantic Segmentation in Aerial Imagery using Wavelet-Enhanced Cost-sensitive Symmetric Fully Convolutional Neural Networks}

\author{Seyed~Majid~Azimi,~
         Peter~Fischer,~
         Marco~K\"orner,~
          and~Peter~Reinartz
\thanks{Seyed Majid Azimi, Peter Fischer, and Peter Reinartz are with the Department 
of Earth Observation Center, Remote Sensing Technology Institute, Photogrammetry and Image Analysis, German Aerospace Center (DLR) , Oberpfaffenhofen, M\"unchenerstra{\ss}e 20, 82234 Wessling, Bavaria, Germany.
Corresponding author e-mail: seyedmajid.azimi@dlr.de.}
\thanks{Seyed Majid Azimi and Marco K\"orner are with Technical University of Munich - Department of Civil, Geo and Environmental Engineering, Chair of Remote Sensing Technology, Arcisstra{\ss}e 21, 80333 Munich, Bavaria, Germany}
}

\maketitle

\begin{abstract}
The knowledge about the placement and appearance of lane markings is a prerequisite for the creation of maps with high precision, necessary for autonomous driving, infrastructure monitoring, lane-wise traffic management, and urban planning.
Lane markings are one of the important components of such maps.
Lane markings convey the rules of roads to drivers.
While these rules are learned by humans, an autonomous driving vehicle should be taught to learn them to localize itself.
Therefore, accurate and reliable lane marking semantic segmentation in the imagery of roads and highways is needed to achieve such goals.
We use airborne imagery which can capture a large area in a short period of time by introducing an aerial lane marking dataset.
In this work, we propose a Symmetric Fully Convolutional Neural Network enhanced by Wavelet Transform in order to automatically carry out lane marking segmentation in aerial imagery.
Due to a heavily unbalanced problem in terms of number of lane marking pixels compared with background pixels, we use a customized loss function as well as a new type of data augmentation step.
We achieve a high accuracy in pixel-wise localization of lane markings \change{compared with the state-of-the-art methods} without using 3rd-party information.
In this work, we introduce the first high-quality dataset used within our experiments which contains a broad range of situations and classes of lane markings representative of today’s transportation systems.
This dataset will be publicly available and hence, it can be used as the benchmark dataset for future algorithms within this domain.
\end{abstract}

\begin{IEEEkeywords}
Lane Marking Segmentation, Fully Convolutional Neural Networks, Wavelet Transform, Infrastructure Monitoring, Traffic Monitoring, Autonomous Driving, Mapping, Remote Sensing, Aerial Imagery.
\end{IEEEkeywords}

\IEEEpeerreviewmaketitle

\section{Introduction}
\IEEEPARstart{N}{owadays,} the detailed description of the public transportation network is essential for the generation of accurate road maps and lane based models.
A broad range of current services, \eg navigation systems and assisted driving rely on such information.
Future applications like automated lane-wise traffic monitoring, urban management and city planning are also asking for high precision maps at centimeter-level accuracy, particularly built for autonomous driving applications which are called \gls{HD} maps.
At present \gls{AV} are a research focus in computer vision and remote sensing.
In order to achieve autonomy in \glspl{AV}, one key factor is to localize the vehicle precisely.
\goal
Very accurate maps containing the location of infrastructures such as streets, sidewalks, traffic lights and even lane markings are a necessity for reaching the goal of fully autonomous driving.
\Gls{ADAS} comprising features like vehicle navigation and lane departure warning requires not only the road model information, but also the precise road lane marking data, \eg the lane marking types and their locations.

\TODO{Traffic Monitoring} Besides the current omnipresent topic of autonomous driving, many more urgent topics can be addressed by \gls{HD} maps.
For instance the traffic monitoring systems could benefit from the localization of lane markings as the base map.
Information about lane marking locations in open-space parking lots could also result in a more complete and therefore more efficient parking lot utilization.
In addition, more applications can arise which will use high precision maps as the smart and efficient management of transportation systems is one of the main topics of the 21st century.

At present the data collection for generating \gls{HD} maps is mainly carried out by so called mobile mapping systems, which comprise in most cases of a vehicle equipped with a broad range of sensors (\eg Radar, Lidar, cameras).
This method comes with some drawbacks, for instance the ground based systems can cover only a small part of the map due to the sensor line-of-sight.
Sensor drift and \gls{GPS}-shadows in urban canyons lower the spatial accuracy, traffic flow leads to partial occlusions in the recorded data.
This issue can be addressed by remote sensing imagery which are intrinsically motivated by the need for large areas in short time at a monetary competitive level.
\challenges
More and more airborne and space-borne sensors recording data in the very-high resolution, \eg \change{\gls{gls:GSD}} less than \SI{50}{\cm} domain are in operational mode.
The public sector often offers its data under a free-and-open policy, \eg aerial imagery of \change{\gls{USGS}} in urban regions has \gls{gls:GSD} less than \SI{30}{\cm}.
Data collected by flight campaign with the goal to monitor infrastructure can offer even better \gls{gls:GSD}.
\cref{fig:goal} gives an example of such imagery from AerialLanes18 dataset, introduced in this work which can be used for the purpose of \gls{HD} maps creation.

\subsection{Challenges}
Several issues raise the level of difficulty when it comes to image segmentation of aerial imagery for creating \gls{HD} maps.
Some of them are well known general problems in the computer vision domain, for instance:

\begin{itemize}
\item Occlusion (partial or full) changes the appearance of lane markings in the image.
Some occlusion cases can be observed in \cref{fig:challenges}: full occlusion can be caused by other objects such as bridge, tree and so on, while
partial occlusion which occurs more often is mostly caused by trees.

\item Shadow creates a different illumination over lane markings causing changes in their appearance.
It does not happen often that lane markings are overshadowed, making it a special case.
This reason, like the previous one, could reduce the accuracy of automatic lane marking algorithms, especially Deep Learning methods which need a lot of training samples.
\classes
\end{itemize}
Some other challenges are specifically binded to the task of lane marking segmentation.
A short overview is given in the following itemization.
\begin{itemize}
\item Different classes - Generally, lane markings are categorized into different classes such as single and double boundary, intersection, boxed junction, separator, zig-zag, special sign for the disabled, bus and bike sign, speed limit, no-parking zone, pedestrian crossing, and so on.
Some of these classes can be seen in \cref{fig:classes}.

\item Small size - In airborne imagery, the size of lane markings compared to other objects in the image is, depending on the GSD, quite small.
In some cases, a sign of separator could be \SI{5x5}{\px}.
This is one of the biggest challenges within the lane marking mapping task in aerial imagery.

\item Washed out samples - Not all lane markings are visible in the image; some of them appear washed out partially or completely.
This imposes another challenge for the accurate localization of lane markings.
On the one hand, in the case of completely washed out lane markings, no visual feature may be captured.
Therefore, these cases are ignored.
Partially occluded objects, on the other hands, impose a difficult challenge both in the prediction and dataset annotation steps.

\item Rare cases - Lane marking classes are not equally distributed, as some classes are more frequent than others.
Speed limit, bus and bike signs, parking place for the disabled can be named as rare cases which can be seen in \cref{fig:challenges}.
\similarobjects
\item The complex background represents an additional hindrance in accurate localization of lane markings.
Structures such as those in \cref{fig:similarobjects} resemble with high similarity lane markings.

\end{itemize}

\subsection{Related work}
Besides of the before mentioned challenges concerning semantic lane-marking segmentation of aerial imagery, another challenge was identified in the early phase of this work.
The usage of aerial images in order to extract valuable data from transportation infrastructure has a rich literature in the remote sensing domain.
But as it comes to supervised learning algorithms, we identified the lack of annotated, high-quality datasets.
As the lane markings are so small, annotating such objects is difficult and time-consuming.
We will later on tackle this issue by making our dataset easily available.

Concerning aerial imagery Jin et al.~\cite{jin:remotesensing2012} firstly extracts the roads.
\change{Then they apply Gabor filters for highlighting the lane markings followed by Otsu's thresholding algorithm for raw binary segmentation.
The final result is then given by morphological operators or by using \glspl{SVM}~\cite{burges1998tutorial}.
However, by using this approach some white linear features such as the ridges of house roofs may be misclassified if the road extraction is not applied.
Also lines belonging to vehicles or bridges may be misclassified as they are inside the road areas.
Furthermore, they did not investigate lane-marking extraction into detail providing only one resulting image.} 
They also mentioned that objects, such as trees above roads or worn-out/dirty lane markings on the roads, decrease the accuracy of the final results.
In order to solve the problem, Jin et al.~\cite{jin:icsps2010} propose an approach consisting of three steps to detect lane markings:
\begin{itemize}
\item First, the road centerline is extracted.
\item Then the road surface is detected.
\item Finally pavement markings are extracted.
\end{itemize}
\change{Similar to the previous work, in this work also roads are extracted first and then lane-markings are detected.
Even though, this method shows better performance than previous methods as claimed by the author, it still has the drawback of the previous methods such as not being able to have a good accuracy on lane-marking detection without road extraction.} 
Following this work flow, Jin et al.~\cite{jin:automaticremotesensing2012} use an unsupervised algorithm to extract the road surface first.
Second the authors employed co-occurrence contrast measurements to enhance the lane markings, under the assumption that the contrast between lane marking and road surface is strong and then localized lane markings.
Subsequently, morphological closings and openings are applied in order to remove the enhanced edges in the shadow regions.
In the last step, the extracted lane marking features are narrowed by a modified Wang-Zangen algorithm and further fitted to a line by least square regression.
\change{This work extends lane-marking detection to rural areas.
Similar to the previous mentioned works, despite yielding good results in the few provided test images, this work also suffers from high rate of false positives in case of not using road extraction step.}
\overview
Further works following this core approach are given by Javanmardi et al.~\cite{javanmardi2017} \change{and Huang et al.~\cite{Huang2014} who used adaptive threshold in airborne images.
Javanmardi et al.~\cite{javanmardi2017} approach contains different steps such as \gls{DSM} processing, removing vehicles using multiple images and in the end utilizing a simple adaptive thresholding to extract lane marking.}
\change{In this method, lane markings are not detected directly as we have done in this work and 3rd party data is used to remove non-lane marking objects.}

Hinz and Baumgartner~\cite{Hinz2003} propose a method to extract lane markings by multi-view imagery and context cues and also used the extracted thin lines as a hint for the presence of a road.
\change{This method yields very good results.
However, this method works only when multiple images have been captured with different views from a place of interest.
This method is also similar to previous mentioned works in using the road mask and therefore it suffers from low accuracy in case of not applying the road extraction step.}
Mattyus et al.~\cite{7780762, gellerticcv17} proposed a method based on Markov Random Fields and a combined parsing of both ground and aerial images to generate detailed maps.
\change{These road models could be used for masking images in order to localize lane markings, but it can not be used directly for lane marking localization and only helps to find roads and the boundaries of each line in the roads.}

Tournaire and Paparoditis~\cite{OlivierTournaire2007} extract dashed-line and zebra crossing with the use of information obtained by the reconstruction process from the extracted primitives of the image.
\change{In contrast with our work, they only considered rectangle line markings and studied their geometric properties to be able to extract them.
Furthermore, they did not use a learning feature approach to detect lane markings as we have done in this work.}
More complete overviews about the extraction of roads and road features from airborne images can be found in Mayer et al.~\cite{Mayer06atest} and Wang et al.~\cite{WANG2016271}.

As discussed, no previous work has tried to learn the features of the lane marking through an end-to-end feature learning mechanism \eg deep learning methods, to the best of knowledge of these authors.
Unlike in remote sensing community, researchers in computer vision community have already applied deep learning methods to extract road infrastructure features in in-situ images.

Deep learning methods, currently widely used in computer vision, try to learn features rather than using engineered features.
During the last few years, deep learning methods have shown impressive performance in a variety of computer vision tasks such as object recognition~\cite{Simonyan2015VeryRecognition, resnetHe15,huang2017densely}, detection~\cite{fasterrcnnNIPS2015,he2017maskrcnn,ssdeccv16,redmon2017yolo9000} and semantic segmentation ~\cite{Long2015FullySegmentation,Chen2014SemanticCRFs, pspcvpr2017,unet}.
\Glspl{gls:CNN}, as one of the widely used deep learning methods, have been proven to be very successful for object recognition in images ~\cite{Simonyan2015VeryRecognition,resnetHe15,huang2017densely}.

However, pixel-wise semantic segmentation is a more challenging problem, as each pixel should be classified.
Kim et al.~\cite{kimcvpr17} propose a sequential transfer learning method based on \glspl{gls:FCNN} by segmenting the road in the first step and then lane marking segmentation on the road-masked image.
\change{This method is similar to the methodology used in current lane-marking detection algorithms in remote sensing.
The main difference is now using \glspl{gls:FCNN} to extract roads first rather than using non-deep-learning-based methods.}

Gurghian et al.~\cite{Gurghian_2016_CVPR_Workshops} propose a \gls{gls:CNN} classification method to localize lane markings on both sides of a vehicle.
\change{However, this method is not applicable to remote sensing applications as we are interested to detect lane-marking in all regions in the images.}
Lee et al.~\cite{LeeKYSBKLHHK17} propose a multi-task \gls{gls:CNN} to localize and classify lane markings in day time with different weather conditions as well as during night time.
\change{This is a very interesting work where the author has developed a method to detect lane-markings in different weather conditions.
However, this method and other \gls{gls:FCNN}-based methods in lane-marking detection have been developed for ground imagery processing.
Lane-markings of small size in image data have not been the focus of most works in this context. In imagery from cars or poles (ground imagery) they are big enough and therefore do not introduce a significant challenge.
Having said that, in remote sensing imagery lane-markings can be as small as \SI{3x3}{\px} which are much more difficult to detect.}

In order to facilitate the application of supervised learning methods, Caltech Lane~\cite{aly08realtime} and tuSimple~\cite{tusimple17} datasets were created for lane marking segmentation, while large-scale datasets for semantic understanding of roads containing a diverse range of classes including lane markings have been defined in ~\cite{Cordts2016Cityscapes,Neuhold_2017_ICCV}.
\change{The aforementioned datasets are in ground imagery and to the best of our knowledge there is no public dataset available for research on lane marking localization in remote sensing data.}

\change{In our work, we have created the first high-quality annotated dataset for lane-marking semantic segmentation in remote sensing imagery specifically in airborne images.
	We use \glspl{gls:FCNN} as baselines of our method.
    Therefore, this work is to our knowledge the first time using \glspl{gls:FCNN} to segment lane-marking in remote sensing data in contrast to previous methods which mostly detect road first as a hint and secondly apply edge detection-based methods to segment lane-markings.
	This is one of the main differences of this work compared to previous works on this task.
	Unlike the works done in ground imagery, in this work we focus on small size lane-markings by inserting \glspl{DWT} of input images in different steps into \glspl{gls:FCNN} to preserve high-frequency information including lane-markings.
	Wavelet transforms have been widely used both in ground~\cite{Hu09-WaveBlood} and remote sensing imagery~\cite{Mallat2009}.
	Recently, Fujieda et al.~\cite{FujiedaTH17} also used \gls{DWT} combined with \glspl{gls:CNN} for texture classification.
	They used \glspl{gls:CNN} for classification while in our work the focus is on the semantic segmentation task which is a different task from classification.
	They inserted all \gls{DWT} decompositions with \gls{gls:CNN} only in two steps and in the middle of the convolutional layers and did not investigate which insertion place for \gls{DWT} yield the best results while in our work we use three decompositions and also we investigate where is the best place to insert DWT to yield the best results.
	In their work, \gls{DWT} decompositions were inserted into \glspl{gls:CNN} as input while in our work, we still give RGB image as input.
    More importantly, the effect of \gls{DWT} was not investigated from the point of preserving high-frequency data such as very small objects for semantic segmentation.
	Moreover, we deploy a weighted loss function as well as symmetric \gls{gls:FCNN}.
	Although, \glspl{gls:FCNN} introduced by Long et al.~\cite{Long2015FullySegmentation} is among the first deep learning methods for the semantic segmentation task, its accuracies are still comparable with the-state-of-the-art such as DeepLabv3~\cite{deeplabv3}, DeepLabv3+~\cite{deeplabv3+}, PSPNet~\cite{pspcvpr2017}, ICNet~\cite{icnet} and others with deep backbone networks such as ResNet~\cite{resnetHe15}, ResNext~\cite{resnext}, Xception~\cite{xception}, and DenseNet~\cite{densenet}.
	We choose the \gls{gls:FCNN} network proposed by Long et al.~\cite{Long2015FullySegmentation} with VGG16 backbone as baseline of our method due to its simplicity and familiarity of the community with its architecture and yet its accuracy is comparable with the-state-of-the-art methods.
}
\subsection{Our contribution}
In this work, we focus on lane marking pixel-wise semantic segmentation using aerial images.
In high-resolution aerial images, the lane markings are easy to identify.
Our proposal is based on combining \glspl{gls:FCNN} with \gls{DWT} for lane marking pixel-wise semantic segmentation in airborne images.
The motivation of using \glspl{gls:FCNN} as a deep learning method for semantic segmentation is its higher performance compared with non-deep-learning methods.

Unlike traditional methods in which feature extraction and classification steps are performed separately, in \glspl{gls:FCNN} features are learned during an end-to-end training and there is no separation between feature extraction and feature classification.
\Glspl{gls:FCNN} have been proposed first by Long et al.\cite{Long2015FullySegmentation} for semantic segmentation in in-situ imagery with extra up-sampling layers (deconvolutional layers).
The authors propose multiple pooling layers to be fused with up-sampling layers (skip layers) to further refine segmentation boundaries.
The authors call their network and its variants FCN32s, FCN16s and FCN8s.
We consider FCN32s as the base-line of this work.

\TODO{With Wavelets}In order to enhance current network performance, we combine different  of input images with the \gls{gls:FCNN} network.
The motivation of using \gls{DWT} is to provide the network with different representations of input objects in different scales as well as full-spectral analysis.
\Glspl{DWT} can represent the input image at different scales.
While \glspl{gls:CNN} process the image in the spatial domain and partially in the spectral domain, DWT allows analyzing the images in full-spectral domain.
Therefore, the properties of these algorithms are different.

Integrating \gls{DWT} will enable the network to access the \change{intensity frequency information} which is lost in the convolution and average pooling layers, carrying out limited spectral analysis.
\change{The intensity frequency information lays in the frequency domain for the pixel intensities variation and not in the different image bands like e.g. in hyperspectral images.}
Wavelet transform has been investigated for a long time for frequency analysis and also image compression.

In this work, we have carried out experiments with different combinations of DWT decompositions to be used as input with a modified version of FCN32s, which we call ``Symmetric FCNN''.
The final result is a pixel-wise semantic segmentation of lane-marking.
Due to the heavily unbalanced task in terms of number of lane marking pixels compared to background ones, we have applied a cost-sensitive loss function to impose higher loss for wrong classification of lane markings as minor class than loss for the wrong classification of background.
As mentioned before we introduce the first high quality pixel-wise annotated dataset for lane marking segmentation and detection in aerial imagery, which shall encourage future works in this area.

The following sections are organized as follows.
\cref{sec:methods} represents the methodology to enhance \gls{gls:FCNN} with different \gls{DWT} decompositions, the cost-sensitive loss function used during the training phase, and the symmetric \glspl{gls:FCNN} architecture.
In \cref{sec:experiments}, we introduce the dataset and its features and properties and report different experiments.
In \cref{sec:results}, the results of the experiments are given and evaluated.
In \cref{sec:conclusion} a conclusion is drawn.

\section{Arial LaneNet: Wavelet-enhanced Cost-sensitive Symmetric Fully Convolutional Neural Network} \label{sec:methods}

In this work, we propose a Cost-sensitive Symmetric \gls{gls:FCNN} enhanced by \gls{DWT} which we call Aerial LaneNet.
The overall work-flow of our method is illustrated in \cref{fig:overview}.
Due to the high resolution of aerial images and hardware memory constraint, the original images are chopped into small patches using a sliding window\cite{dalal}.
Then each patch is processed by Aerial LaneNet in order to predict a semantic segmentation of the input patch.

The output is a binary image which denotes which pixel belongs to lane markings and which one to the background.
In the end, patches are stitched together to create the final output with the same resolution as the input image.
In the following, we explain our proposed methods in detail.

\Glspl{gls:CNN} are a combination of different layers such as convolution, pooling, activation function, drop-out and fully connected layers.
Input data is convolved with a linear convolution filter in convolution layers
\begin{align}\label{eq:convolution}
(h_{k})_{ij} = (W_{k}\ast X)_{ij} + b_{k}
\end{align}
where $k=1,\dots,K$ is the $k$-th feature map in the convolution layer and $(i, j)$ is the index of a neuron in it.
$X$ stands for the input data and $W_{k}$ and $b_{k}$ are the weights (trainable parameters) of the network and the biases (trainable parameters) respectively.

The output of each neuron in the $k-th$ feature map, has been represented by $(h_{k})_{ij}$ at position $(i, j)$.
The 2D convolution between input data and filter mask in spatial domain is represented by $``\ast"$ which partially includes spectral analysis at low-frequencies, while the remaining spectral information is lost.

Considering Figure \ref{fig:wavelet1}, parts shown in red in the \gls{DWT} algorithm can be considered as a convolution function in traditional \glspl{gls:CNN}.
On the other hand, a wavelet transform is able to capture the full spectral information of the input in the frequency domain.

Moreover, wavelets can extract multi-resolution spectral information from input data at different decomposition levels as shown in \cref{fig:wavelet2}.
Multi-resolution analysis of the input data would represent the input in different scales, similarly to a pooling operation.
Each sub-sampling step in wavelet transform can be considered as a different pooling operation.

Therefore, pooling layers could be also replaced by wavelet transforms.
Instead of doing so, we merge (fuse) wavelet information of the input with traditional \glspl{gls:FCNN} together with pooling layers which can be done in different ways.
\waveletone    
In order to add the wavelet decomposition to the network, one can compute wavelet transforms for each image and apply the output to \glspl{gls:FCNN}.
However, in this case, multi-scale information of the data is lost.
Therefore, the network is not able to learn the lane marking features at different resolutions.
This will lead to a non-scale-invariant method.
To address this problem, multi-scale input processing is needed.

Each level of wavelet decomposition analyzes the data at different resolution.
Therefore, by combining different decomposition levels of wavelet transforms with \glspl{gls:FCNN}, low and high frequency domain analyses as well as different resolution analysis are achieved.

After applying a wavelet transform on the input image, lane marking boundaries appear as high-frequency objects in vertical, horizontal and partially in diagonal details in the wavelet coefficients.
Different parts from the first to the third level of the \gls{DWT} are illustrated in \cref{fig:wavelet2}.
\wavelettwo
\change{\subsection{Discrete Wavelet Transform (Background)}
	\Gls{DWT} of a signal $x$ is computed by applying a series of filters and sub-sampling in subsequent levels~\cite{Mallat2009}.
	For instance, in the first level of \gls{DWT}, a low pass and a high pass filter are applied simultaneously with impulse responses of $g$ and $h$ resulting in two convolutions of 
\begin{align}\label{eq:dwt}
	\begin{split}
	y_{lowpass}[n]&= (x*g)[n] = \sum_{x=-\infty}^{+\infty}x[k]g[n-k] \\
	y_{highpass}[n]&= (x*h)[n] = \sum_{x=-\infty}^{+\infty}x[k]h[n-k]
	\end{split}
	\end{align}
and the resulting signals are sub-sampled by a factor of 2 \ie
	\begin{align}\label{eq:dwt}
	\begin{split}
	y_{lowpass}&= (x*g) \downarrow 2\\
	y_{highpass}&= (x*h) \downarrow 2
	\end{split}
	\end{align}
	In order to further increase the approximation coefficients and the frequency resolution resulting from low and high pass filters and down-samplings, this decomposition is repeated.
	This results in a tree representation of each decomposition level known as filter bank which is illustrated for a two-level decomposition in \cref{fig:wavelet1}.
	We can consider the implementation of a wavelet filters as the wavelet coefficients calculation of a discrete set of lower-level wavelets for a mother wavelet function$\Psi(x)$.
    By applying \gls{DWT}, a discrete function $f(x)$ is converted into a signal of two variables~\cite{Mallat2009}: scale and translation which can be described as
	\begin{align}\label{eq:dwt-sc}
	\Psi_{j,k}(x)&:= \dfrac{1}{2^{j/2}}\Psi(\dfrac{x - k2^{j}}{2^j})\\
	\Phi_{j,k}(x)&:= \dfrac{1}{2^{j/2}}\Phi(\dfrac{x - k2^{j}}{2^j})\\
	\Psi(x)&:=  
	\begin{cases}
	1				& \text{,for }  0\leq x\leq 1/2\\
	-1				& \text{,for } 1/2< x\leq 1\\
	0               & \text{,otherwise}
	\end{cases}\\
	\Phi(x)&:=  
	\begin{cases}
	1& \text{,for }  0\leq x\leq 1\\
	0              & \text{,otherwise}
	\end{cases}
	\end{align}
	in which ${\Phi_{j,k}(x)}$ is the scaling function for which the box function $\Phi$ has been chosen.
	$\Psi_{j,k}(x)$ and ${\Phi_{j,k}(x)}$ have ranges of $[-\dfrac{1}{2^{j/2}}, \dfrac{1}{2^{j/2}}]$ and $[0, \dfrac{1}{2^{j/2}}]$ accordingly with width $2^j$ that starts at $k2^j$.
	The scale level is represented by $j$ and the shift by $k$.
	$\Psi_{j,k}(x)$ are scaled and shifted versions of the continuous mother wavelet $\Psi(x)$.
	In the discrete domain, for a signal of length $N = 2^n$ one considers the
	$N$ functions $\Phi_{n,0}, \Psi_{n,0} \dots \Psi_{1,2^{n-1}-1}$.
	In this work, we consider the Haar wavelet transform as the first order of the Daubechies wavelet family~\cite{daubechies1992ten} with $n=2$ and we use the basis vectors
	\begin{align}
	\begin{split}
	\Phi_{2,0}&= \frac{1}{2} (1, 1, 1, 1)^T \\
	\Phi_{2,0}&= \frac{1}{2} (1, 1, -1, -1)^T\\
	\Phi_{1,0}&= \frac{1}{2} (1, -1, 0, 0)^T\\
	\Phi_{1,1}&= \frac{1}{2} (0, 0, 1, -1)^T
	\end{split}
	\end{align}
	that yield the coefficients
	\begin{align}
	\begin{split}
	c_{j,k}&:= f^T\Phi_{j,k} \\
	d_{j,k}&:= f^T\Psi_{j,k}
	\end{split}
	\end{align}
	in which $c_{j,k}$ are coefficients of the scaling vector $\Phi_{j,k}$, for coarse decomposition these are low-pass filter coefficients.
    Similarly $d_{j,k}$ are coefficients of the wavelet vector $\Psi_{j,k}$ for detailed decompositions which are high-pass filter coefficients.
	In 2D \gls{DWT}, it starts first with calculating the wavelet decomposition on a single level in $x$ direction then in $y$ direction.
	Afterwards the next decomposition is performed only in the quadrant part that contains the low-frequency parts (scaling coefficients) for both directions.
	The decomposition levels are proceeded until a single pixel is reached.}

In order to compress the images as wavelet transform injections, the orthonormal Daubechies wavelet family~\cite{daubechies1992ten} is selected for their proven success in decomposing images and identifying borders.
The Daubechies wavelet family is written as dbN, where N is the order, and db is the abbreviation for the Daubechies wavelet family.
The db1 wavelet is the same as the Haar wavelet \change{and the first order of Daubechies family with lower computation cost and fewer wavelet filter bank coefficients.
The continuous wavelet transform has been presented in \cref{eq:dwt-sc}}.
	
As shown in \cref{fig:overview}, DWT decompositions are injected as shown by the paths in pink.
Given that the input data is H(Height) and W (Weight) pixels \change{after having changed to gray-scale image shown in Figure \ref{fig:wavelet2}}, using four levels of the wavelet transform on the input image results in the outputs with $H/2\times W/2$, $H/4\times W/4$, $H/8\times W/8$ and $H/16\times W/16$ sizes.
\change{The input image is first converted to gray-scale before DWT computation.
In contrast to usual cases in which more data results into a better performance, our preliminary results show that using an RGB input image results in 1.78\% \gls{IoU} performance decrease.
To further investigate this issue, we considered other color spaces including HSV and observed the same effect which we conjecture it could be due to insertion of redundant input data.
It is worth mentioning that the parameters of \gls{DWT} is fixed and are not updated during the training phase.}
The first level \gls{DWT} has an input size of H$\times$W, and four outputs (Approximate, Horizontal, Vertical, and Diagonal) with half size capturing different details in the image like shown in \cref{fig:wavelet1}.

The fusion of the 1st level wavelet transform has to be done after the first pooling.
The reason is that the input size of the image is H$\times$W while the size of the 1st level wavelet decomposition is H/2$\times$W/2.
Hence, due to incompatible size resolution, the first fusion layer is carried out after the first pooling operation.

Inserting the 1st level \gls{DWT} decompositions with half size of the input image as input to the network results in losing spatial and spectral information of the original input.
Therefore, this scenario is not efficient.

There are different ways of wavelet transform fusion with the FCNN network, as shown in \cref{fig:differentfusion}.
As mentioned, the wavelet decompositions have to be placed after the pooling layer.
We have considered all three illustrated cases to combine the 1\st wavelet decomposition level to the network.
The same goes for other \gls{DWT} levels.
\differentfusion    
A typical cross entropy loss function in semantic segmentation treats pixels belonging to different classes equally.
For a binary classification problem, this can be represented as
\begin{align}\label{eq:crossentropy}
\resizebox{0.5\textwidth}{!}{$L(\mathbf{W}) = -\frac{1}{N}\sum_{n=1}^{N}{y_n\log{\hat{y}(x_n,\mathbf{W})}+(1-y_n)(1-\log{\hat{y}(x_n,\mathbf{W}))}}$}
\end{align}
where $x_n \in [0,255]$ is the input pixel value, $y_n \in \{0,1\}$ the ground truth label, $\hat{y}_n \in [0,1]$ the prediction probability, $\mathbf{W}$ is the weight matrix of the network and $L$ denotes the loss function.

In order to classify each pixel, the softmax function is widely used in multi-class classification tasks in \glspl{gls:FCNN}.
The vector of real values between $[0, 1]$ generated by this function denotes a categorical probability distribution.

The softmax function can be expressed as $\hat{y}_j= softmax(X,W_j) = \frac{e^{X^{T}W_j}}{\sum_{k=1}^{K}e^{X^{T}W_j}}$, in which $W_j$ and $X$ denote the weights of the network (including bias values) and the input data respectively.
The well-known loss layer using the softmax function for multi-class classification is cross-entropy loss.

However, for lane marking segmentation, the majority of pixels belong to the non-lane marking class.
This makes the problem highly unbalanced.
Therefore, we modify the typical cross entropy loss function by imposing a higher cost on the wrong classification of a lane marking pixel compared with a background pixel.
The defined loss function is
\begin{dmath}\label{eq:weightedcrossentropy}
L(\mathbf{W})= -\frac{1}{N}\Big(\lambda_{lane}\sum_{n=1}^{N}{y_n\log{\hat{y}(x_n,\mathbf{W})}}+\sum_{n=1}^{N}{(1-{y}_n)\log{(1-\hat{y}(x_n,\mathbf{W}))}}\Big)
\end{dmath} 
which is cost-sensitive, as it penalizes different class pixels differently.
This is done by introducing parameter $\lambda_{lane}$ in the cross entropy loss function.
\change{This weighted loss function can be easily extended to a multi-class segmentation scenario by inserting a function $\mathds{1}_{cls} (x_n)$ which is equal to one if $x_n$ belongs to class $cls$ and zero if it does not.}
To leverage the capacity of \glspl{gls:CNN} to perform semantic segmentation, the networks can be modified by replacing fully-connected layers with convolution layers which allow \glspl{gls:CNN} to be applied to images with variable sizes.

This approach will not lead to semantic segmentation with the same resolution as the input image.
Therefore, extra up-sampling layers (bi-linear interpolation) are applied in the base-line network.
Bi-linear interpolation is differentiable which makes applying back-propagation during training feasible.

In order to grasp varied visual input information yet keeping input feature map dimensions, the up-sampling layer is applied after the last convolution layer to up-sample the extracted features to the input dimension size.
This can be considered as encoding of the input data to the first up-sampling layer and decoding by up-sampling layers as illustrated in \cref{fig:overview}

By modification of \glspl{gls:FCNN} to be more robust to over-fitting, we design a symmetric \gls{gls:FCNN} network.
In this methodology, we add convolution and \change{drop-out} layers after up-sampling layers in the baseline network of FCN32s.
We do the same for FCN16s and FCN8s network architectures.
We also add one additional up-sampling layer which can be seen as a new FCN4s network.

\change{Instead of using average pooling layers, we use max-pooling layers.
In FCN4s, we also apply the fusion technique used in the baseline paper which is a summation of the corresponding pooling layers with the output of the up-sampling layers.}
\change{The motivation to add more convolution layers comes from ~\cite{Simonyan2015VeryRecognition,Krizhevsky2012ImageNetNetworks,resnetHe15}  where it has been shown that depth has a key role in high-level feature extraction.}

Aerial LaneNet is not limited to a fixed input size \ie there is no need to resize input images.
The only preprocessing step is the subtraction of image mean.
Due to the heavily unbalanced datasets for lane marking and the scarcity of such datasets, more drop-out layers have been added to the network to prevent over-fitting.
The deep neural networks are prone to over-fitting according to the noise present in the training set samples if that is small.

The inserted layers have been denoted in red in \cref{tab:receptivefields}.
In \cref{fig:breakdown}, the Aerial LaneNet network architecture is reported in detail.
\breakdown
In order to investigate the architecture of the network and its properties such as input and output size, feature map dimension, receptive field and so on, \cref{tab:receptivefields} has been prepared.
\receptivefieldsTable
\section{Experiments} \label{sec:experiments}
In this section, we introduce the dataset used in the experiments.
Then we explain the experiments and provide quantitative and qualitative results along with corresponding discussions.

\subsection{AerialLanes18 Dataset}
The experiments were conducted using images acquired by the German Aerospace Center (DLR) within a flight campaign in the \change{framework} of the VABENE++ project.
The campaign was carried out over the greater area of the city of Munich on the 26\th of April 2012.

The 3K camera system~\cite{dlr89701} consisting of three Canon Eos 1Ds Mark III cameras was used for recording the raw data, where two cameras are mounted side looking and one is mounted nadir-looking on a flexible platform.

The 3K system is a low-cost camera system used for various remote sensing applications, such as real-time mapping~\cite{kurz2012low}, disaster monitoring~\cite{kurz2012low2}, traffic monitoring~\cite{Liu_Mattyus2015}, and detection of high-density crowds~\cite{meynberg2016detection}.

In total, 20 representative RGB images of size \SI{5616x3744}{\px} have been chosen.
The flight height of about \SI{1000}{\m} above ground led to a \gls{gls:GSD} of approximately \SI{13}{\cm}.

The images depict urban and partly rural areas with highways and first/second order roads.
Complex traffic situations like crossings and congestions are included.
The images served as starting point for works in the domain of vehicle detection by Liu and Mattyus~\cite{Liu_Mattyus2015}.

\change{\subsection{Annotation of AerialLanes18}
The ground truth has been annotated by human experts who marked all kinds of lane markings over roads and highways such as separate line, continuous line, turn sign, speed limit sign, and even bus and disabled people parking place signs.
The annotation was carried out manually by using an in-house annotation software.
During annotation, we ignored washed out lane-markings.
}
\cref{fig:patchtrimg1} shows some patches of the mentioned dataset.
\cref{fig:wholetrimg1} show large training images with the overlaid lane marking annotations.
\patchtrimgone
\wholetrimgone
\subsection{Implementation Details}
As the dataset does not consist of many images, most likely training a deep neural network on such a small dataset from scratch with randomly initialized parameters will lead to over-fitting.
On the other hand, as annotating small lane marking objects is difficult and time-consuming, only images of the mentioned dataset have been annotated.
\change{To address this problem, networks which have already been trained using large datasets like ImageNet~\cite{imagenet} are used as initialization of parameters in order to transfer the learned information to a new task.
This technique is known as ``Transfer Learning''.}
Using this technique, we can initialize the weights more efficiently.

\change{Therefore, it can be assumed that the network is already close to one of the optimal solutions and needs far less training data to converge and by retraining the network known as ``Fine-tuning'' technique, the problem of over-fitting can decrease significantly.}
In our experiments with wavelet transform fusion, we use FCN32s~\cite{Long2015FullySegmentation} as the baseline.
VGG16 proposed by Simonyan et al.~\cite{Simonyan2015VeryRecognition} is the backbone main network.
However, AlexNet~\cite{Krizhevsky2012ImageNetNetworks}, GoogleNet~\cite{szegedy14googlenet}, and ResNet-101~\cite{resnetHe15} are also considered.

\change{We use the patches of \SI{1024x1024}{\px} as input to the network.
We employ the \SI{800}{\px} cropping step in horizontal and vertical directional in the training phase and \SI{1000}{\px} in the test phase.
For the training step, random flipping patches are applied for data augmentation.}
We consider one random image as validation set which consists of 24 patches.
In the test set, the number of test patches is 240.
Networks are trained on the training set to find the best hyper parameters and then \change{both the} training plus the validation set are used for the final training.

It should be mentioned that in the following experiments no extra information such as road segmentation or third-party data such as OpenStreetMap~\cite{openstreetmap} has been used.

Aerial LaneNet is trained end-to-end.
The optimization problem of finding the minimum value in the loss function is solved by Adam optimizer~\cite{Kingma2015Adam:Optimization} and Back-propagation~\cite{lenet} process.
\change{The learning rate of 0.0001 with batch size of 1 is used.
We have trained the final network for about 10 epochs on one Nvidia Titan X Pascal GPU using the Tensorflow~\cite{tensorflow2015} framework.}

\section{Results and Evaluation} \label{sec:results}
In our experiments, we compare the final output of the system for each image (not patch) with the corresponding ground truth.
Therefore, in lane marking segmentation, the goal is to classify each pixel as lane marking class (foreground) or non-lane marking (background).
The more pixels are classified correctly, the more accurate the system is.
Concerning the evaluation criteria, we use the metrics used by  Long et al.~\cite{Long2015FullySegmentation} which are widely used in semantic segmentation tasks.
In these metrics, $n_{ij}$ is the pixel number belonging to class $i$ which has been predicted as class $j$ and $n_{cl}$ stands for the number of classes with $t_i=\sum_{j}n_{ij}$ representing the total number of pixels belonging to class $i$.
IoU means intersection over union \ie it is proportional to the intersection between predictions and ground truth.
\2
\lambdaTable

We use the dice similarity coefficient also due to the heavy unbalance in the dataset.
The number of pixels belonging to each class does not have effect on these two criteria.
$P$ and $T$ represent prediction and ground truth respectively.
The criteria are derived as follows:
\begin{itemize}
    \item Pixel accuracy: \begin{align}\label{eq:pixacc} \frac{\sum_{i}n_{i,i}}{\sum_{i}t_{i}} \end{align}
    \item Mean accuracy: \begin{align}\label{eq:meanacc} \frac{1}{n_{cl}}\sum_{i}\frac{n_{i,i}}{t_i} \end{align}
    \item Mean IoU: \begin{align}\label{eq:meaniu} \frac{1}{n_{cl}}\sum_{i}\frac{n_{i,i}}{t_i+\sum_{j}n_{j,i}-n_{i,i}} \end{align}
    \item Frequency weighted IoU: \begin{align}\label{eq:fmeaniu} (\sum_{k}t_{k})^{-1}\change{\sum_{i}} \frac{t_{i}n_{i,i}}{t_i+\sum_{j}n_{j,i}-n_{i,i}} \end{align}
    \item Dice similarity coefficient: \begin{align}\label{eq:fmeaniu} \frac{2\mid P \cap T \mid }{\mid P  \mid +\mid T \mid} \end{align}
\end{itemize}
\change{and recall and precision are calculated using the criteria
\begin{align}
\begin{split}
Recall &:= \frac{True\, Positives}{True\, Positives + False\, Negatives}\\
Precision &:= \frac{True\, Positives}{True\, Positives + False\, Positives}~.
\end{split}
\end{align}
}
The baseline network of FCN32s with AlexNet as backbone network is trained from scratch and due to the small and highly unbalanced dataset, it classifies lane-marking pixels as background in most areas, with only 51.0\% mean \gls{IoU} accuracy.

Employing weighted loss increased the performance by almost 2 percent by penalizing wrong classification of lane marking pixels more than wrong classification of background pixels, alleviating to some extend the challenge posed by an unbalanced dataset.
\lambdaEffect

Before applying the customized loss function, fine tuning using a pre-trained model trained on ImageNet~\cite{imagenet} as well as data augmentation are applied, due to the small training dataset available.
\subsubsection{\change{Different Base Network Investigation}}
Results in \cref{tab:2} show the performance of Aerial LaneNet in lane marking segmentation with different network architectures.
VGG16 outperforms AlexNet as the shallower network and slightly GoogleNet.
The high pixel accuracy of this system should be investigated as most of pixels belong to the background class rather than lane markings.
\change{This phenomenon has two main reasons: firstly the network is over-fitting to the background class due to the small-size dataset and secondly due to the heavily unbalanced dataset.}
\symmetric
As expected, due to the highly unbalanced dataset, pixel accuracy and frequency weighted IoU are larger than 99\%.
These parameters, as mentioned before, are not suitable to evaluate performance of a network using a highly unbalanced task.
That is the why mean \gls{IoU} and Dice are more reliable criteria to evaluate an algorithm in such cases.
\eachimageone
\eachimagetwo

\subsubsection{\change{The Effect of $\lambda$}}
The value of $\lambda_{lane}$, which is a hyper-parameter, should be tuned.
There is no automatic approach to find the best value for this parameter.
One approach is considering the default value of $\lambda_{lane}=389$ as the ratio between background to lane marking pixels in the training set.
Another method is grid search which can be applied to refine the default value.
We considered the pixel ratio in the test set as well as other setups ranging from 1 to 1000.
With this approach, we noticed that the pixel ratio is not the best value to get the best results (\cref{fig:lambda}).
Considering \cref{tab:lambda}, the best value is achieved with 400 which is higher than the default one and lower than 418 as the ratio in tainval set.
Performance degrades using 308 as the ratio in the test set.
This shows the network has learned this hyper-parameter based on the training set.
In this case, more research can be devoted to find the best value of $\lambda_{lane}$ automatically.\\

\subsubsection{\change{The Importance of Symmetric \gls{gls:FCNN}}}
As mentioned in the last section, in order to extract higher-level features as well as making the network robust to noise in the training set, a symmetric \gls{gls:FCNN} is designed.
The improvement introduced by this algorithm shown in \cref{tab:symmetric} is almost 3 percent in terms of mean \gls{IoU}.
Adding more convolution, drop-out and up-sampling layers seem to have almost the same impact of around 1 percent point on the mean \gls{IoU}.
This indicates that even though deeper network could basically improve the performance, the major problem is not their depth.
\5
An observation of symmetric \gls{gls:FCNN} networks shows that even if the network is deep, the algorithm has some difficulty to segment small lane markings.
Due to the nature of low-frequency spectral analysis of \gls{gls:FCNN}, lane markings are smoothed and removed after convolution and average pooling operations.
To address this problem, wavelet transform of input image is inserted into the network.

\subsubsection{\change{The Effect of \gls{DWT}}}
Multi-resolution analysis using different levels of wavelet transform augments the performance by considering lane marking objects at different scales.
\6
\cref{tab:5} indicates that a combination of the first four \gls{DWT} decomposition levels results in the best performance, confirming our motivation for multi-resolution analysis.
In our experiments we noticed that the addition of a 5th level worsens the results, which could be due to small size lane markings, since most of their details have already been discarded.

\change{In order to further improve the performance, we replaced the VGG16 base network with the ResNet-101~\cite{resnetHe15} network which has better performance on the ImageNet dataset in comparison to VGG16. 
We inserted \gls{DWT} levels after the first pooling layer in stage 1 and after the first convolution layer with stride of 2 in each stage from stage 2 to stage 4.
We did not insert \gls{DWT}'s 5\nnth level to stage 5 due to our observation in the \gls{DWT}'s 5\nnth level insertion after the last pooling layer in VGG16 (cf.\ \cref{tab:5}).}

As wavelet transform decomposition is made of horizontal, vertical, diagonal details as well as an approximation component, investigation is carried out to investigate the effect of each component.
\sFigure

\subsubsection{\change{The Effect of \gls{DWT} Components}}
According to Table \ref{tab:6}, horizontal and vertical components have considerably more impact than the other two.
Although the diagonal component also increases mean \gls{IoU} by almost 2 percent points, it has less effect than the rather horizontal and vertical components of almost 5 percent.
This indicates that the majority of lane markings are present in the horizontal and the vertical \gls{DWT} components.
The approximation part, however, worsens the performance.
This could be due to the fact that this part does not carry sparse information about lane marking as other parts.
Experiments with orders of Daubeschies wavelet transforms higher than 1 resulted in lower performance of ~1.45 mean \gls{IoU} for db2 which could be due to  less appearance of the lane marking in higher Daubeschies orders.

\subsubsection{\change{Varied Possible Fusions}}
As shown in \cref{fig:differentfusion}, Table \ref{tab:7} reports the result of different \gls{DWT} fusion with symmetric \gls{gls:FCNN}.
We have considered three different fusion locations.
The fusion can be either after the pooling layers or convolution layer or before the pooling layers.
Before the first pooling layer, due to dimension incompatibility, the fusion is not possible.
Results in \cref{tab:7} show that placing the fusion right after the pooling layers results in the best performance.
The reason for this phenomenon could be the extraction of high-level features by subsequent convolution layers.
In contrary, fusion of DWT decomposition before pooling layers leads to a decrease in mean \gls{IoU}.
This could be due to the reason that \gls{DWT} representation is pooled by the next pooling layer which smooths the representation.
However, this degradation is not significant, as lane marking pixels have higher values compared to neighboring pixels, and in max pooling operation the maximum value is chosen.

\7

\subsubsection{\change{Confusion Matrix Investigation}}
In order to evaluate true and false positives/negatives in our method as well as precision and recall, we have considered the confusion matrix of the configuration for the best performance.
\cref{tab:confusionmatrix} indicates, that in spite of a heavily unbalanced dataset, the system is able to achieve a lane marking pixel (pixel-wise) accuracy of 71.55\%.
\3
\BenchmarkFigure
In spite of different illumination conditions introduced by shadows, different shapes and sizes, the network is able to classify background pixels with 0.1\% false positive compared with 99.8\% true negative pixels.
This indicates how robust the system is in the presence of the very complex background and objects similar to lane-marking.
However, the false negatives are still high.

The majority of false negative cases come from straight and dot-shape lane markings.
In straight lane markings, the output width of the system is almost in all of cases narrower than ground truth.
This indicates this architecture is not able to segment boundaries accurately.
Although a morphological operation could increase the performance in this case dramatically, it is not interesting from a research point of view and we do encourage other researchers not to use it in next researches on this dataset for benchmarking.

As mentioned, dot-shape objects yield a considerable number of false negatives.
These objects are as small as \SI{5x5}{\px} which makes them difficult to segment.
However, as we do not have access to the information of which pixel belongs to which class in the current annotation, we cannot report a number in this case.

Another and important source of false negative is shadows.
As shadows occur rarely, the network has not been able to learn shadows to segment lane markings accordingly.
Regarding rare objects, like "BUS" signs, speed limits, disabled parking places, turn signs and so on, the same phenomenon is happening.
These classes do not occur often and as in deep convolutional neural networks a big number of training samples is needed to train the network, performance in these cases is not high.
\confusionmatrixTable
\subsubsection{\change{Comparison with the state-of-the-art}}
\change{We also compared Aerial LaneNet with FCN-8s, DeepLab~\cite{Chen2014SemanticCRFs}, UNet~\cite{unet} and the state-of-the-art method DeepLabv3~\cite{deeplabv3}~\footnote{\label{note1}\url{https://github.com/tensorflow/models/tree/master/research/deeplab}}, and its newer version DeepLabv3+~\cite{deeplabv3+}~\footref{note1} in \cref{tab:3}.
Interestingly, there is a big gap between DeepLabv3+ and DeepLabv3.
The reason is that DeepLabv3 uses monotonically increasing atrous rates which in spite of being effective to obtain large receptive field to segment large-size objects, it severely damages information from small objects like lane markings.
In contrast, DeepLabv3+ uses a multi-scale encoder containing atrous convolutions to obtain a multi-scale contextual information and in the decoder part a simple yet effective module refines the segmentation outputs to improve the boundary segmentation.
The qualitative comparison has been provided in \cref{fig:qualitativebenchmark}.
The multi-scale processing helps the DeepLabv3+ to achieve significantly better results than its previous version.
This is mostly due to the decoder part which improve the boundary region segmentation.
However, it does not have a satisfactory performance on tiny lanemarkings despite its very good performance in the terrestrial images.
The results shows that recovering high-frequency information of image pixels by inserting DWT into different levels of \glspl{gls:CNN} leads to a considerably better performance of 4\% mIOU in comparison with DeepLabv3+ algorithm.
Aerial LaneNet outperforms all of these networks in \cref{tab:3} showing the high accuracy of our method.}

\subsubsection{\change{Qualitative Analysis}}
In \cref{fig:eachimage1}, recall and precision values for each test image are reported.
These values are consistent and there is not a big difference between recall and precision.
In \cref{fig:eachimage2} mean \gls{IoU} and Dice for each test image as well as recall and precision for each class have been reported.
As for total recall and precision values, these criteria are consistent among test images.
Recall and precision values for each class have also been computed.

One can notice that precision and recall for background class is very high, which is due to the unbalanced task: there is a big gap between recall and precision for the lane marking class and for the background class.
In order to evaluate the results qualitatively, \cref{fig:s} illustrates the lane marking segmentations of different patches of size \SI{1024x1024}{\px} compared with the ground truth.
The left images are input test patches.
The middle patches are the ground truth.
The patches on the right are the corresponding predictions.
These figures show a very good performance in the segmentation of both straight and dashed lines in highways.
It is very interesting that in some cases the network has localized correct lane marking which are not even annotated in the ground truth.
\overlaidorgone
\newtestpatch
However, there are also some failure cases.
In the same figures, one can note that shadows, narrower straight lines, very small lane markings, and similar objects in the background are the main reasons for false negative and positive outputs.
\cref{fig:s}(a) shows the shadow caused by a truck has caused degradation in lane marking segmentation.
Objects with similar appearance still are a challenge \eg the roof structures at the left bottom part of image in \cref{fig:s}(b), which look similar to lane markings have been classified as lane marking.
Also in the same image, when it comes to smaller lane marking objects, the network is not performing as good.
\newtestpatchoverlaid
In spite of these failure cases, the overall performance proves the concept of effective semantic segmentation of lane marking using enhanced \glspl{gls:FCNN} with \gls{DWT} information.
In \cref{fig:overlaidorg1} predictions have been overlaid on the original test images after stitching prediction patches together.
In these images, predicted lane marking pixels and undetected ones are reported in red and blue respectively.
In shadow areas the network has difficulties to segment lane markings.

\subsubsection{\change{Cross-domain Generalization}}
In order to evaluate the robustness of our algorithm to variations: GSD, camera angle view, and illumination conditions, we have considered \change{multiple flights on different days}, altitudes and angles with the DLR 3K camera.
Results are reported in \cref{fig:newtestpatch}.

We have over-laid predictions on test patches of a new dataset in \cref{fig:newtestpatchoverlaid}.
The performance shows a good generalization capability of the network, which appears robust to most of the challenges mentioned earlier such as small size, different camera angles and presence of objects similar to lane marking such as lanes in soccer fields.

\section{Conclusions} \label{sec:conclusion}
In this work, we have introduced a reliable and fast algorithm to segment very small objects such as lane markings in aerial imagery with high accuracy and robustness.
We presented the Aerial LaneNet network based on the idea of enhancing \glspl{gls:FCNN} with wavelet transformation coefficients for pixel-wise semantic segmentation, which enables a full spectral and multi-scale analysis resulting in the considerable improvement compared with our FCNN based-line network.
\change{We have shown that using sub-sampling layers or atrous convolutions  to obtain large receptive fields although yields very good performance in terrestrial images, they cause a vital data lost for pixel-wise semantic segmentation of tiny objects which leads to a considerable performance degradation.
Therefore, the lost information should be either injected into the network or be kept by removing sub-sampling layers to recover the lost data.
In this work, we selected the first strategy showing impressive performance improvement in comparison with the state-of-the-art methods.
We conclude that for tiny object segmentation both high and low frequency information of pixels should be analyzed while \glspl{gls:CNN} perform mostly low frequency analysis due to using pooling and convolution layers.
The limitations of Aerial LaneNet is in shadow areas, semantic signs on the roads as well as washed out lane-markings.}
We also introduced the AerialLanes18 dataset the first high-quality aerial lane marking dataset as a benchmark in this domain.
Using different levels of wavelet decomposition leads to a multi-resolution data analysis which is important in extracting lane markings, as objects appear at different scales.
In the future, we will investigate improving the performance by processing shadow areas differently.

\bibliographystyle{IEEEtran}
\bibliography{bibliography}
\end{document}